\documentclass[11pt]{article}
\usepackage[preprint]{acl}

\usepackage{times}
\usepackage{latexsym}
\usepackage[T1]{fontenc}
\usepackage[utf8]{inputenc}
\usepackage{microtype}
\usepackage{inconsolata}

\usepackage{graphicx}
\usepackage{placeins}

\usepackage{amsmath}
\usepackage{amssymb}

\usepackage{booktabs}
\usepackage{makecell}
\usepackage{subcaption}
\usepackage{xcolor}
\usepackage{alltt}
\usepackage{array}
\usepackage{colortbl}
\usepackage{makecell}
\usepackage{pgf}
\usepackage{caption}
\usepackage{tabularray}
\UseTblrLibrary{booktabs}
\definecolor{sigLightGreen}{RGB}{220,242,220}
\definecolor{rhoPosBlue}{RGB}{0,92,175}
\definecolor{rhoNegRed}{RGB}{180,35,24}

\usepackage{siunitx}

\usepackage{fontawesome5}

\usepackage{algorithm}
\usepackage{algpseudocode}

\PassOptionsToPackage{most}{tcolorbox}
\usepackage{tcolorbox}

\usepackage{listings}

\usepackage{hyperref}
\hypersetup{
    colorlinks=true,
    linkcolor=blue,
    citecolor=blue,
    urlcolor=blue
}

\usepackage{cuted}

\tcbset{
  colback=white,
  colframe=black,
  boxrule=0.3pt,
  arc=2pt,
  left=4pt,
  right=4pt,
  top=4pt,
  bottom=4pt,
}

\newcommand{\xhdr}[1]{\par\medskip\noindent\textbf{#1}\par\smallskip}

\definecolor{headerblue}{RGB}{220,230,242}
\definecolor{sigGreen}{RGB}{0,180,90}
\newcommand{\heat}[2]{%
  \colorbox{yellow!#1}{\makebox[3.8em][c]{\strut #2}}%
}
\definecolor{sumcolor}{RGB}{220,240,255}
\definecolor{qacolor}{RGB}{220,255,220}
\definecolor{mathcolor}{RGB}{255,235,220}

\definecolor{prfblue}{RGB}{0, 70, 140}     %
\definecolor{prforange}{RGB}{180, 90, 0}   %
\definecolor{prfred}{RGB}{160, 30, 50}     %

\newcommand{\prf}[3]{%
  \textcolor{prfblue}{#1}\hspace{0.6em}%
  \textcolor{prforange}{#2}\hspace{0.6em}%
  \textcolor{prfred}{#3}%
}

\newtcblisting{qualpromptbox}{
  width=\linewidth,
  colback=gray!5,
  colframe=gray!60,
  boxrule=0.5pt,
  arc=2pt,
  left=6pt,
  right=6pt,
  top=6pt,
  bottom=6pt,
  listing only,
  breakable,
  enhanced,
  listing options={
    basicstyle=\ttfamily\small,
    breaklines=true,
    breakatwhitespace=true,
    breakindent=0pt,
    columns=fullflexible,
    keepspaces=true
  }
}

\newtcblisting{qualoutputbox}[1]{
  width=\linewidth,
  colback=white,
  colframe=gray!60,
  boxrule=0.5pt,
  arc=2pt,
  left=6pt,
  right=6pt,
  top=4pt,
  bottom=4pt,
  breakable,
  enhanced,
  title=\textbf{#1},
  fonttitle=\normalsize,
  listing only,
  listing options={
    basicstyle=\ttfamily\small,
    breaklines=true,
    breakatwhitespace=true,
    breakindent=0pt,
    columns=fullflexible,
    keepspaces=true
  }
}

\newenvironment{qualexample}[4]
{
\begin{tcolorbox}[
  colback=white,
  colframe=black,
  boxrule=0.8pt,
  arc=2pt,
  breakable,
  enhanced,
  width=\linewidth,
  title=\textbf{#1},
  fonttitle=\normalsize,
  before skip=8pt,
  after skip=8pt
]
{\small\color{black!70}
#2 \hspace{1em} #3 \hspace{1em} #4\par}
\vspace{0.4em}
}
{
\end{tcolorbox}
}
\title{
DepthKV: Layer-Dependent KV Cache Pruning \\
for Long-Context LLM Inference
}

\author{
Zahra Dehghanighobadi$^{1,2}$ \quad
Asja Fischer$^{1}$ \quad \\
$^{1}$Ruhr University Bochum \\
$^{2}$UAR Research Center for Trustworthy Data Science and Security \\
\small Correspondence: 
\href{mailto:zdehghanighobadi1998@gmail.com}{zdehghanighobadi1998@gmail.com},
\href{mailto:asja.fischer@rub.de}{asja.fischer@rub.de}
}

\date{}

\begin{document}

\maketitle

\begin{abstract}
Long-context reasoning is a critical capability of large language models (LLMs), enabling applications such as long-document understanding, summarization, and code generation. However, efficient autoregressive inference relies on the key–value (KV) cache, whose memory footprint grows linearly with sequence length, leading to a major memory bottleneck. To mitigate this overhead, KV cache pruning methods discard cached tokens with low attention scores during inference. Most existing methods apply a uniform pruning ratio across layers, implicitly assuming that all layers contribute equally to overall model performance. We show that this assumption is suboptimal, as layers differ significantly in their sensitivity to pruning. We propose \textbf{DepthKV}, a layer-dependent pruning framework that allocates a fixed global KV budget across layers based on their sensitivity, rather than using a uniform allocation. Across multiple models and tasks, DepthKV consistently outperforms uniform pruning at the same global pruning ratio, demonstrating more effective utilization of the KV cache budget through layer-dependent allocation.\\
\\
\faGithub\ \href{https://github.com/Zahra1998Dehghani/depthKV}{github.com/zahra-dehghani/depthkv}
\end{abstract}

\section{Introduction}
\begin{figure*}[!t]
    \centering
    \includegraphics[width=\linewidth]{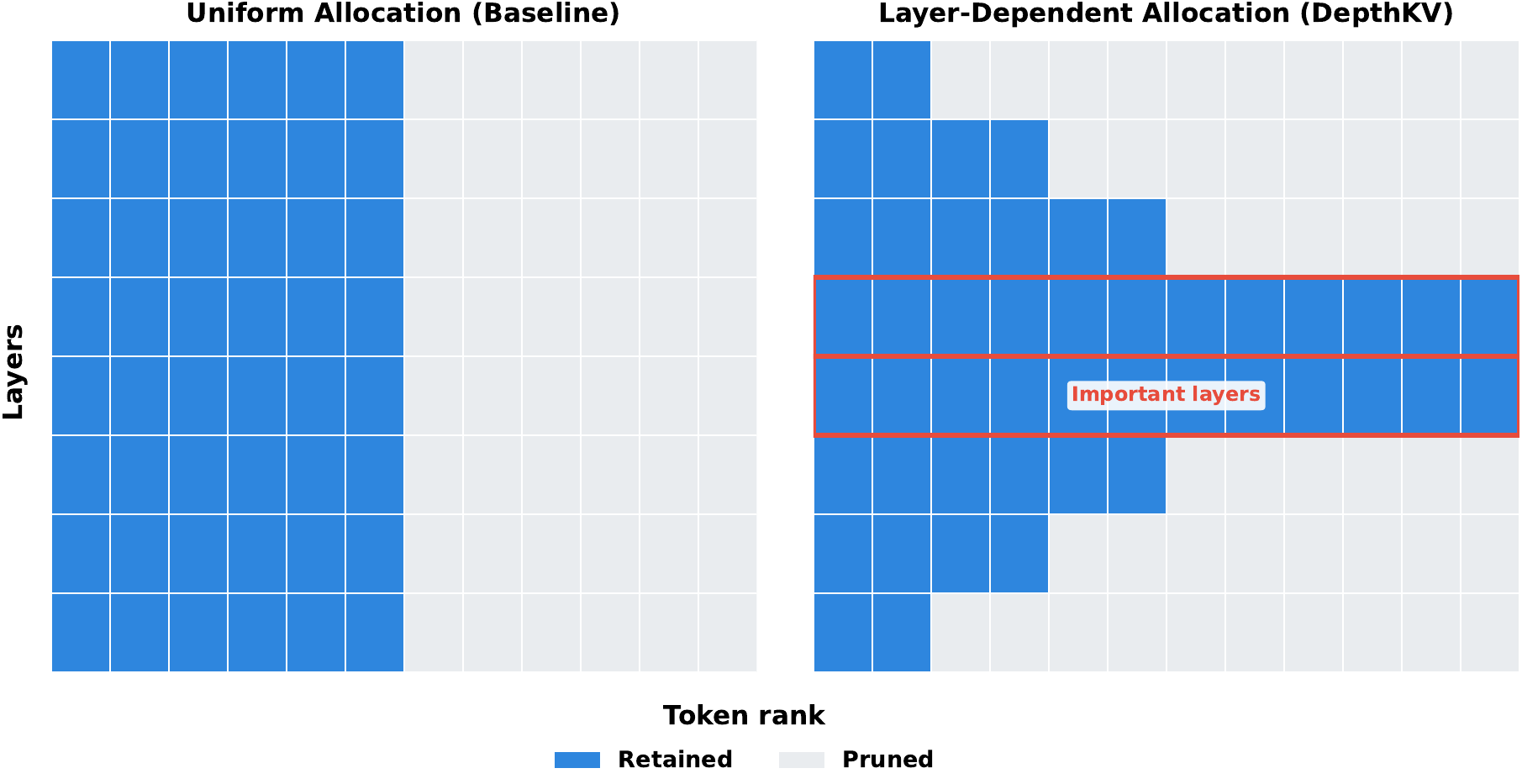}
    \caption{
    \textbf{Uniform vs.~layer-dependent KV allocation.}
    Uniform allocation (left) assigns an equal KV budget across transformer layers.
    DepthKV (right) reallocates this budget based on sensitivity to pruning, retaining more tokens in critical layers (highlighted) and pruning less important ones more aggressively.
    Token rank denotes relative importance.
    }
    \label{fig:depthkv_overview}
\end{figure*}

Recent advances in large language models (LLMs) have greatly increased context window sizes, ranging from 128K to millions of tokens \citep{team2024gemini, glm2024chatglm}.
Larger context windows enable applications such as long-form reasoning, agent-based workflows, and large-scale document retrieval, where relevant information is often sparsely distributed across lengthy inputs.%

However, in autoregressive LLMs, each generated token attends to all previous tokens via self-attention. Decoding scales linearly with context length and is repeated at every generation step, while the prefill stage processes the entire input with quadratic complexity. Therefore, as context windows grow, long-context inference becomes prohibitively expensive.

To avoid redundant computation, modern LLMs cache key–value (KV) representations of previous tokens. While this eliminates recomputation during decoding, it introduces a new bottleneck: the KV cache grows with both sequence length, quickly exceeding GPU memory capacity in long-context settings \citep{wu2024retrieval}. As a result, the primary bottleneck shifts from computation to memory.

Prior work addresses this challenge at different stages of the model lifecycle. Training-phase approaches alter the attention architecture but typically require retraining \citep{shazeer2019fast, ainslie2023gqa, brandon2024reducing}. Deployment-stage methods focus on optimizing how the KV cache is stored and accessed at the system level, such as memory layout and hardware placement, without changing its values or the model’s computations \citep{kwon2023efficient, lin2024infinite, ye2024chunkattention}. In contrast, post-training approaches directly modify the KV cache representation, for example through eviction, merging, or quantization, and often introduce approximations to improve efficiency \citep{zhang2023h2o, wang2024model, yang2024no, hooper2024kvquant}.

Training-phase methods are difficult to apply to existing pretrained models, while deployment-stage approaches mainly optimize memory access rather than reducing KV cache size. In contrast, post-training methods directly reduce memory usage during inference, making them particularly practical for long-context settings \citep{shi2024keep}. Therefore, we focus on post-training KV cache pruning.

Most existing post-training methods prune the KV cache uniformly across transformer layers, implicitly assuming that all layers are equally important. However, prior work \citep{skean2025layer} suggests that intermediate transformer layers may play a more critical role than early or late layers. To examine whether such non-uniformity persists under KV cache pruning, we conduct a layer-wise ablation study in which pruning is applied to one layer at a time while keeping others unchanged, and measure the resulting performance degradation. A permutation test consistently rejects the hypothesis of uniform layer importance across models and datasets, demonstrating that transformer layers contribute unevenly to long-context performance.

We further analyze how layer removal affects generation behavior. As shown in Section~\ref{subsec:content_amplification}, layers that are most sensitive in the ablation study also lead to shorter and less informative outputs when pruned, indicating that their impact on performance is closely tied to their role in sustaining content generation.

Motivated by these findings, we propose \textbf{DepthKV}, a framework for layer-dependent KV cache pruning that allocates the memory budget non-uniformly across transformer layers based on their importance for long-context performance. As illustrated in Figure~\ref{fig:depthkv_overview}, uniform allocation retains the same number of tokens across all layers, whereas DepthKV assigns a larger budget to more important layers while pruning less critical ones more aggressively, thereby preserving key information under a fixed memory constraint. The framework supports multiple allocation strategies, including position-dependent protection (preserving specific regions such as middle layers), metric-guided allocation (allocating more budget to layers ranked higher by a scoring metric), and hybrid strategies that combine these rules within a unified budget.

We evaluate DepthKV on long-context tasks including document summarization, question answering, and mathematical reasoning, where inputs are substantially longer than outputs, making the prefill stage the dominant source of computation and memory usage. In this setting, pruning during prefill directly targets the primary bottleneck while maintaining stable decoding. Across all tasks and models, layer-dependent allocation consistently outperforms uniform pruning under the same KV budget.

In summary, our main contributions are as follows:

\begin{itemize}
    \item We show that transformer layers exhibit statistically significant variation in importance for long-context inference, challenging the implicit assumption underlying uniform KV cache pruning.

    \item We identify \emph{content amplification layers}, whose pruning suppresses content generation and strongly correlates with downstream performance degradation.

    \item We propose \textbf{DepthKV}, a layer-dependent KV cache allocation framework that redistributes a fixed memory budget across transformer layers based on their importance.

    \item We demonstrate that DepthKV consistently outperforms uniform KV cache pruning across diverse long-context tasks under the same memory budget.
\end{itemize}

\section{Related Work}

Building on the discussion in the introduction, we focus on post-training KV cache pruning methods. Within this setting, existing approaches can be categorized along two complementary dimensions: (i) how token importance is estimated, and (ii) whether this importance depends on the current decoding query. The first distinguishes between heuristic and learned methods, while the second separates query-aware from non-query-aware strategies.

\paragraph{Heuristic methods.}
Heuristic approaches estimate token importance using predefined rules, typically based on positional bias, aggregated attention statistics, or attention profiling across heads or tokens. Representative methods include $H_2O$ \citep{zhang2023h2o}, StreamingLLM \citep{xiao2023efficient}, SnapKV \citep{li2024snapkv}, and FastGen \citep{ge2023model}. $H_2O$ retains attention-dominant tokens alongside recent context.
StreamingLLM preserves initial attention-sink tokens (i.e., tokens that consistently attract high attention, such as the first token \texttt{<bos>}), together with a sliding window of recent tokens. SnapKV selects prefix tokens based on attention patterns computed from an observation window near the end of the prompt, while FastGen derives head-specific retention policies from attention profiling to determine which tokens to preserve. While these methods are computationally efficient, their reliance on heuristic rules may limit generalization under shifts in the input distribution.

\paragraph{Learned methods.}
In contrast, learned approaches estimate token importance directly from data rather than predefined rules, enabling them to better capture semantic and task-specific relevance. Representative examples include DuoAttention \citep{xiao2024duoattention} and SeerAttention \citep{gao2024seerattention}. DuoAttention learns per-head gates to separate retrieval heads with full attention from streaming heads with restricted attention. SeerAttention instead learns input-adaptive block-level sparse attention patterns through a lightweight gating module trained with self-distillation. While generally more flexible, they require an additional training stage and introduce computational overhead.

\paragraph{Query-aware methods.}
Beyond the above distinction, another key dimension is whether token importance depends on the current decoding query. Most existing KV cache pruning methods are non-query-aware, assigning token importance once when tokens are inserted into the KV cache and reusing it throughout decoding. While computationally efficient, such strategies rely on historical information or current states to decide which tokens to discard, making these decisions effectively irreversible. As a result, tokens that appear unimportant at early stages may later become critical for future decoding steps, leading to the loss of relevant information.

In contrast, a separate line of work focuses on query-aware strategies. Query-aware methods dynamically estimate importance at each decoding step, enabling adaptive retrieval but at the cost of additional computation. Representative approaches include Quest \citep{tang2024quest}, RetrievalAttention \citep{liu2024retrievalattention}, and MorphKV \citep{ghadia2025dialogue}. Quest partitions the KV cache into fixed-size pages (i.e., groups of tokens) and uses the current query together with page-level key summaries to estimate the most relevant pages. RetrievalAttention selects the most relevant KV entries for the current query using vector search over indexed keys. MorphKV iteratively updates a fixed-size KV cache based on recent attention patterns.

\paragraph{Positioning of DepthKV.}
Within the above taxonomy, DepthKV can be viewed as a non-query-aware, heuristic post-training method, but differs from prior work by focusing on layer-wise sensitivity instead of token-level importance. DepthKV allocates the KV cache budget across layers based on this sensitivity, achieving a superior memory–performance trade-off and consistently outperforming uniform pruning under the same global pruning ratio.

\section{KV cache pruning}
\label{sec:kv_pruning}

In this section, we formalize the KV cache pruning problem under a global memory budget and briefly discuss previously proposed strategies for uniform KV pruning.

\subsection{Problem Formulation}

We consider a decoder-only transformer with $L$ layers and hidden dimension $d$ in a long-context inference setting.
Given an input sequence of length $N$, each layer produces key–value tensors
$K^{(l)}, V^{(l)} \in \mathbb{R}^{N \times d}$.
Storing all KV pairs requires $\mathcal{O}(LNd)$ memory, which becomes prohibitive for long contexts.

Let $S^{(l)} \subseteq \{1,\dots,N\}$ with $|S^{(l)}| = B^{(l)}$ denote the set of token indices retained at layer $l$, where $B^{(l)} \le N$ is the KV budget allocated to layer $l$. 
The resulting KV memory footprint is proportional to
\[
\sum_{l=1}^{L} B^{(l)} d \enspace.
\]
To ensure a fair comparison across pruning strategies, we impose a fixed global KV budget
\[
\sum_{l=1}^{L} B^{(l)} = B_{\text{total}} \enspace.
\]
The KV cache pruning problem is to select token subsets $S^{(l)}$ and allocate layer budgets $B^{(l)}$ under this constraint while maintaining task performance.

\subsection{Attention-Based Token Importance Estimation}

Pruning is performed during the prefill stage, where token importance is estimated from attention weights over the full input sequence.
Let $Q_i^{(l)} \in \mathbb{R}^{d_k}$ be the query vector of token $i$, and $K_j^{(l)}, V_j^{(l)} \in \mathbb{R}^{d_k}$ the key and value vectors of token $j$ at layer $l$, where $j \le i$.
The scaled dot-product attention score is then given by
\begin{equation}
a_{i,j}^{(l)} =
\frac{\langle Q_i^{(l)}, K_j^{(l)} \rangle}{\sqrt{d_k}} \enspace,
\end{equation}
and the normalized attention weights are
\begin{equation}
\alpha_{i,j}^{(l)} =
\frac{\exp(a_{i,j}^{(l)})}
{\sum_{t \le i} \exp(a_{i,t}^{(l)})}
\enspace.
\end{equation}

\paragraph{H$_2$O w/o V (attention-only).}
Following H$_2$O (Heavy-Hitter Oracle) \citep{zhang2023h2o}, token importance is defined as the cumulative attention assigned to a token by later tokens. 
The importance of token $j$ at layer $l$ is computed as
\begin{equation}
s_j^{(l)} = \sum_{i=j+1}^{N} \alpha_{i,j}^{(l)} \enspace.
\label{eq:h2o_importance}
\end{equation}
In the multi-head setting, attention weights are first aggregated across heads before computing importance scores.
Tokens with the highest importance scores are retained in the KV cache according to the layer-specific budget $B^{(l)}$.

\paragraph{H$_2$O w/ V (value-aware).}
Following \citet{guo2024attention}, token importance is computed by weighting the accumulated attention by the magnitude of the corresponding value vectors:
\begin{equation}
s_j^{(l)} = \|V_j^{(l)}\|_p \sum_{i=j+1}^{N} \alpha_{i,j}^{(l)} \enspace,
\end{equation}
where $\|\cdot\|_p$ denotes the vector norm. 
We consider both $p=1$ and $p=2$, using the former (i.e. the $\ell_1$ norm) by default unless otherwise specified. 
This formulation assigns higher importance to tokens that are both highly attended and associated with large value magnitudes.

\section{Pre-study: Layer-wise Sensitivity to KV Cache Pruning}
\label{subsec:prestudy-kv-pruning-sensitivity}

Prior work \citep{skean2025layer} suggests that intermediate transformer layers may play a more critical role than early or late layers.
Motivated by this observation, we investigate whether this non-uniform importance persists under KV cache pruning across various models and datasets (see Section \ref{sec:experiments} for details on the experimental setup).
To study this, we conduct experiments in which, during the prefill stage, pruning is applied to one layer at a time while all other layers remain unchanged. In particular, for each layer $l$, we apply H$_2$O w/o V and record the resulting performance. Repeating this procedure across all layers yields a layer-wise sensitivity profile.

\begin{figure}[t]
\centering
\includegraphics[width=\columnwidth]{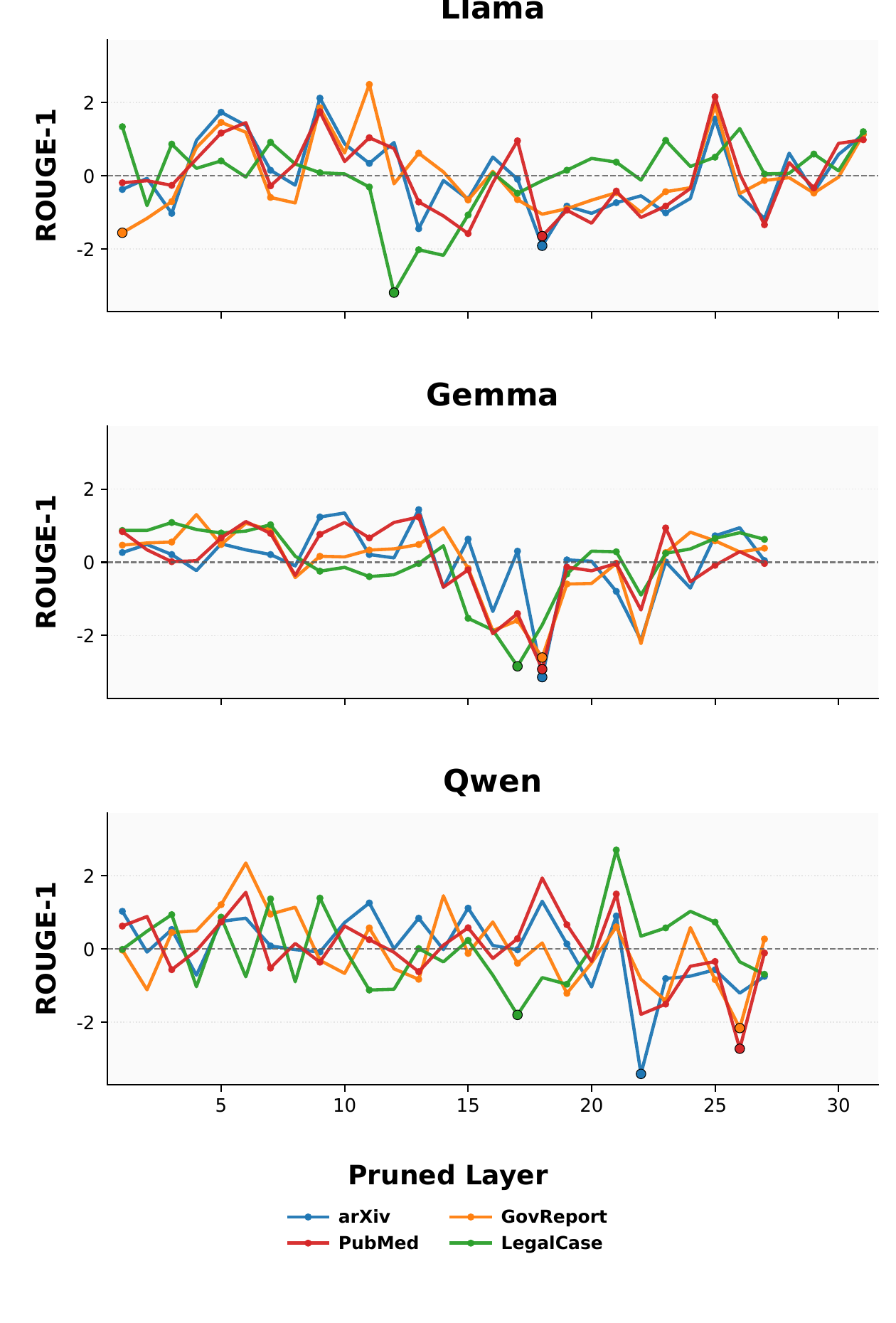}
\caption{
\textbf{Single-layer KV cache pruning.}
Layer-wise ROUGE-1 under KV cache pruning of individual layers, standardized within each model–dataset pair (z-score; mean = 0, standard deviation = 1).
Markers indicate the layer with the largest performance drop for each dataset.
}
\label{fig:layer_ablation_all_models}
\end{figure}

As shown in Figure~\ref{fig:layer_ablation_all_models}, the impact of pruning is highly layer-dependent, with sharp performance drops concentrated at specific layers.
These sensitivity peaks vary across datasets and models, indicating that critical layers are not consistently aligned across settings.
Some layers also exhibit near-zero or positive deviations, suggesting partial redundancy.
This variation underscores the non-uniform, dataset-dependent nature of layer sensitivity.

To quantify this variation, we perform a permutation test on the layer-wise performance differences.
The null hypothesis of uniform layer importance is rejected (permutation test, p-value $< 0.05$; see Appendix~\ref{app:perm_test}), confirming that the observed differences between layers are statistically significant. However, the magnitude of these differences (i.e., the effect size) depends on the dataset and the model.

Overall, these results demonstrate that transformer layers differ substantially in their sensitivity to KV cache pruning, motivating layer-dependent KV allocation strategies.

\subsection{Content Amplification Effects of Layer Pruning}
\label{subsec:content_amplification}

\begin{figure}[t]
\centering
\includegraphics[width=\linewidth]{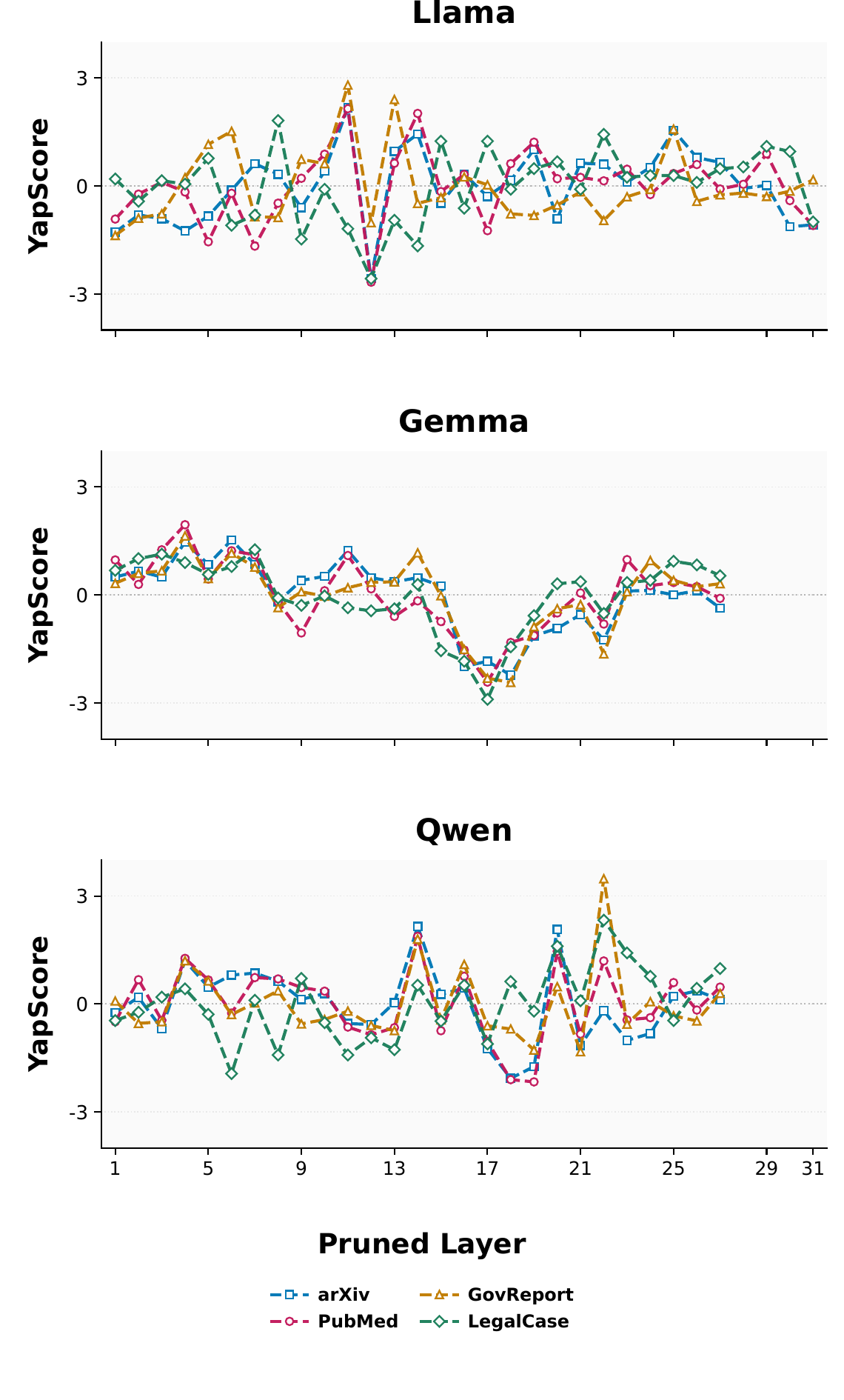}
\caption{
\textbf{Layer-wise normalized YapScore.}
YapScore under single-layer pruning is z-score normalized per dataset. Each curve represents a dataset.
}
\label{fig:yap_layerwise_all_model}
\end{figure}

In this section, we analyze how layer pruning affects generation behavior, revealing an additional dimension of non-uniform layer importance. In particular, we find that pruning certain layers can suppress content generation, leading to shorter or incomplete outputs that degrade summary quality.
Figure~\ref{fig:yap_layerwise_all_model} illustrates the layer-wise YapScore when each layer is independently pruned during the prefill stage across multiple models.

Despite differences across models and datasets, a consistent trend emerges: layers that cause larger reductions in YapScore align with those identified as most sensitive in the pre-study (Section~\ref{subsec:prestudy-kv-pruning-sensitivity}). This suggests that certain layers play a key role in sustaining informative generation, rather than merely influencing output length. We refer to these as \emph{content amplification layers}.

To quantify this relationship, Table~\ref{tab:yap_rouge_corr_longdoc} reports the layer-wise correlation between YapScore and ROUGE-1. Across all models, we observe strong and statistically significant positive correlations, indicating that suppressed content generation is closely associated with performance degradation. We provide qualitative examples supporting these findings in Appendix~\ref{app:qual_examples}.

\begin{table}[t]
\centering
\small
\setlength{\tabcolsep}{3.5pt}
\renewcommand{\arraystretch}{1.12}
\begin{tabular}{lcc|cc}
\toprule
& \multicolumn{2}{c|}{GovReport} & \multicolumn{2}{c}{LegalCase} \\
\cmidrule(r){2-3} \cmidrule(l){4-5}
Model & $r$ & $p$ & $r$ & $p$ \\
\midrule
GEM7  & 0.5202 & $2.49{\times}10^{-4}$  & 0.9902 & $2.89{\times}10^{-39}$ \\
LAM8  & 0.9150 & $3.78{\times}10^{-22}$ & 0.7269 & $4.86{\times}10^{-10}$ \\
QWEN7 & 0.7295 & $8.90{\times}10^{-9}$  & 0.7920 & $5.57{\times}10^{-11}$ \\
\bottomrule
\end{tabular}
\caption{
\textbf{ROUGE-1–YapScore Correlation.}
Pearson correlation coefficients ($r$) and corresponding \textit{p}-values between ROUGE-1 and YapScore, computed across layers under KV cache pruning.
}
\label{tab:yap_rouge_corr_longdoc}
\end{table}

\subsection{Representation Metrics for Layer Importance}
\label{sec:representation_metrics}

To better understand the layer-wise variation observed, %
we analyze representation properties using metrics inspired by \citet{skean2025layer}, which characterize hidden-layer representations in terms of information, geometry, and invariance.
We consider all six of their suggested metrics capturing spectral, geometric, and robustness properties: spectral entropy, effective rank, curvature, DiME, LiDAR, and InfoNCE. Among these, we describe InfoNCE in detail below, as it plays a central role in our subsequent correlation analysis; the remaining metrics are described in Appendix~\ref{app:rep_metrics}.

\paragraph{InfoNCE.} The InfoNCE objective is computed as follows. For each input sequence, we extract a representation matrix $Z \in \mathbb{R}^{T \times d}$ from the post-attention stage of each layer, where $T$ denotes the sequence length, $d$ the hidden dimension, and $z_i$ the representation of the $i$-th token.
The InfoNCE objective measures how well each representation ($i = 1, \dots, T$) remains invariant to input perturbations while staying distinct from other samples in the batch.
To this end, 
first all representations are $\ell_2$-normalized.  We  then construct positive pairs $(\bar z^{(o)}_i, \bar z^{(a)}_i)$, corresponding to the normalized representations of the original and the perturbed input, where each input is perturbed by dropping 10\% of its words uniformly at random.
Representations of other inputs in the batch at the same layer and stage, $\{\bar z^{(o)}_j : j \neq i\}$, serve as negative examples.
The InfoNCE loss for sample $i$ is then defined as:
\begin{equation}
\mathcal{L}_{\mathrm{InfoNCE}}^{(i)}
=
-\log
\frac{
\exp\!\left(\mathrm{sim}(\bar z^{(o)}_i, \bar z^{(a)}_i)/\tau\right)
}{
\sum_{j}
\exp\!\left(\mathrm{sim}(\bar z^{(o)}_i, \bar z^{(o)}_j)/\tau\right)
},
\end{equation}
where $\mathrm{sim}(u,v) = u^\top v$ denotes cosine similarity, and $\tau$ is a temperature parameter controlling the softmax.

\paragraph{Correlation Analysis.}
We evaluate InfoNCE and the other metrics proposed by \citet{skean2025layer} as proxies for layer importance by measuring their correlation with performance degradation under KV cache pruning. Specifically, for each layer, we compute these metrics at four stages of the transformer block---pre-attention, post-attention, post-attention residual, and post-MLP---and assess their correlation with the layer-wise performance drop observed in the pre-study, including statistical significance.
The complete set of correlation results is reported in Appendix~\ref{app:all_correlations}.
Overall, representation metrics frequently exhibit statistically significant correlations with performance degradation, indicating a strong association between representation properties and layer importance under KV cache pruning. Among these metrics, InfoNCE achieves the highest number of statistically significant correlations across settings, with the strongest correlations observed at the post-attention stage, and emerges as the most consistent predictor of layer importance.

\begin{figure}[!tb]
\centering
\includegraphics[width=\columnwidth]{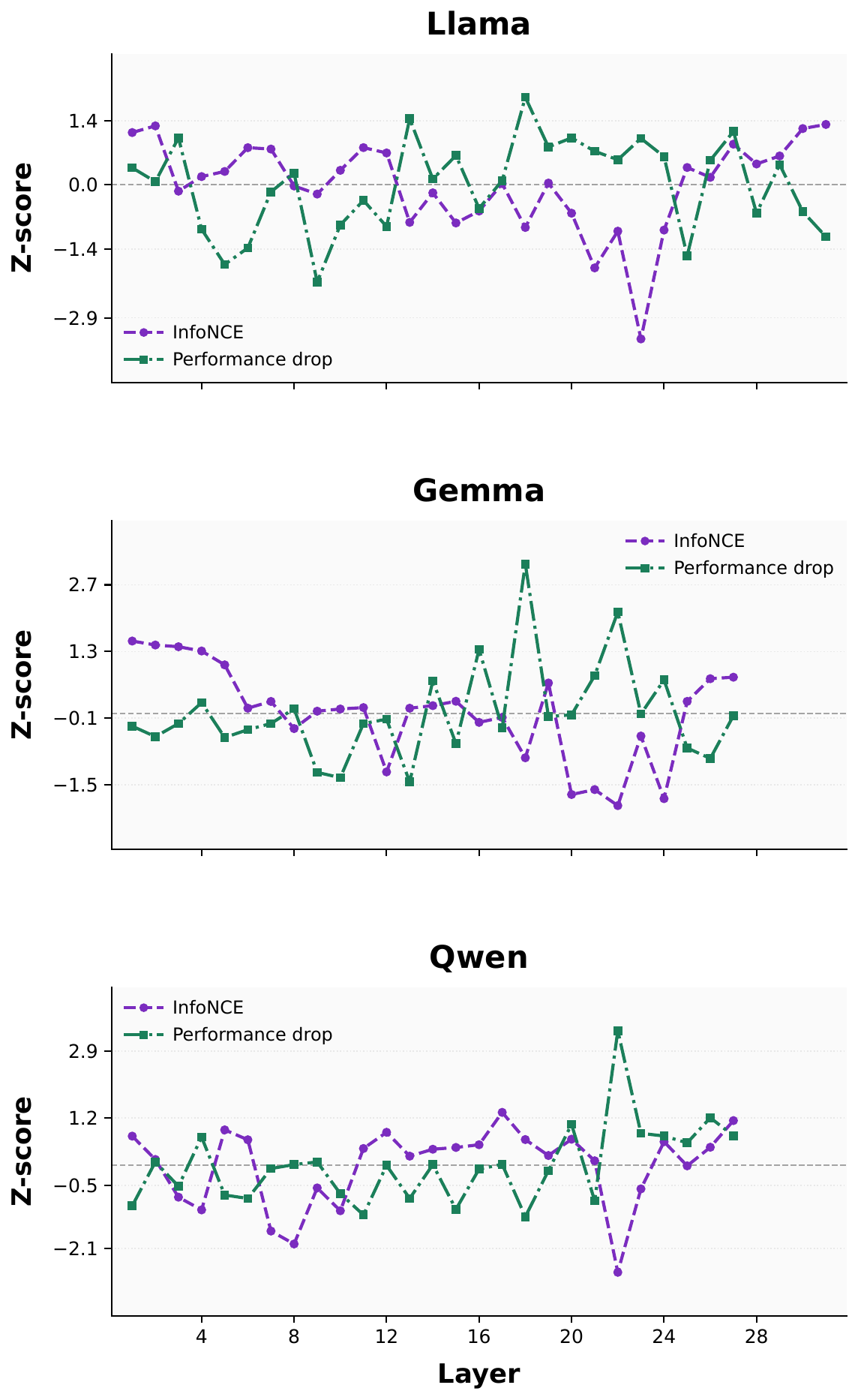}
\caption{
\textbf{Layer Importance vs. InfoNCE.}
Standardized InfoNCE (post-attention) and ROUGE-1 performance drop across layers under KV cache pruning on the arXiv dataset.
}
\label{fig:geometric_vs_drop_arxiv_all_models}
\end{figure}

Figure~\ref{fig:geometric_vs_drop_arxiv_all_models} shows this consistent negative correlation between InfoNCE and performance degradation across models: layers with lower InfoNCE values exhibit larger performance drops when pruned. This indicates that layers whose representations are more robust to perturbations (i.e., lower InfoNCE) are more critical for generation.
This inverse relationship between  InfoNCE and degradation curves across layers suggests that InfoNCE effectively captures depth-wise variation in layer importance.

\section{DepthKV: Layer-Dependent KV Allocation}

Our pre-study reveals that sensitivity to KV cache pruning varies significantly across layers, and further analysis shows that this variation can be predicted from representation-level properties. Building on these insights, we propose \textbf{DepthKV}, a framework that allocates KV budgets across layers according to their relative importance under a fixed global memory constraint.
Let $\rho^{(l)}$ denote the pruning ratio at layer $l$. We impose the constraint
\[
\frac{1}{L} \sum_{l=1}^{L} \rho^{(l)} = \rho,
\]
which ensures that the overall KV budget remains fixed while allowing non-uniform allocation across layers. We do not prune the first layer, in order to preserve the integrity of initial token representations.
Under this framework, we consider three complementary allocation strategies:

\paragraph{Middle-Layer Protection (MLP).}
Motivated by prior findings that intermediate layers play a critical role \citep{skean2025layer}, and further supported by our preliminary analysis, we preserve a subset of middle layers while pruning the remaining layers uniformly. Specifically, we define the middle layers as those surrounding the network midpoint, namely layers $\lfloor L/2 \rfloor$ and $\lfloor L/2 \rfloor + 1$.

\paragraph{Metric-Guided Allocation (MGA).}
We allocate KV budgets according to layer importance scores derived from the InfoNCE metric. 
Since the metric is inversely correlated with performance degradation, we transform it into scores $s^{(l)}$ such that higher values indicate more robust layers. 
We then normalize the scores over the pruned layers as
\[
\alpha^{(l)} = \frac{s^{(l)}}{\sum_{j \in \mathcal{P}} s^{(j)}},
\]
where $\mathcal{P} = \{1, \dots, L-1\}$ denotes the set of pruned layers. 
Pruning ratios are assigned proportionally to $\alpha^{(l)}$, while capping each layer by $\rho_{\max}=0.7$ to avoid overly aggressive pruning. 
Any remaining mass is then iteratively redistributed among unsaturated layers so that the overall allocation satisfies $\sum_l \rho^{(l)} = L\rho$. 
This yields a heterogeneous allocation where more robust layers can tolerate higher pruning, allowing more sensitive layers to retain larger KV budgets.

\paragraph{Middle-Layer Metric Allocation (MLMA).}
We combine structural and metric-based allocation by preserving a subset of middle layers while distributing the remaining KV budget across the other layers using InfoNCE-based importance scores.
We consider three variants, MLMA\text{-}2L, MLMA\text{-}4L, and MLMA\text{-}6L, preserving 2, 4, and 6 middle layers, respectively.

\section{Experimens}
\label{sec:experiments}

We evaluate DepthKV on long-document summarization, document-grounded question answering (QA), and mathematical reasoning tasks, covering diverse domains and long-context reasoning settings.

\subsection{Datasets}
\label{sec:datasets}
 
We consider four long-document summarization benchmarks spanning scientific, biomedical, legal, and government domains (arXiv, PubMed, GovReport, and LegalCase), along with two document-grounded QA benchmarks (Qasper and HotpotQA), and a synthetic mathematical reasoning benchmark (\text{GSM-}$\infty$). 
Table~\ref{tab:input_stats} summarizes key statistics for all datasets, including input length (mean $\pm$ std), task type, and the range of input lengths.
\begin{table}[t]
\centering
\small
\setlength{\tabcolsep}{5pt}
\renewcommand{\arraystretch}{1.15}
\begin{tabular}{l l c c c}
\toprule
\textbf{Dataset} & \textbf{Task} & \textbf{Avg $\pm$ Std} & \textbf{Min} & \textbf{Max} \\
\midrule
\rowcolor{sumcolor}
arXiv      & Sum.  & $4926.7 \pm 1064.2$ & 3053 & 7396 \\
\rowcolor{sumcolor}
GovReport  & Sum.  & $5797.2 \pm 1177.0$ & 3841 & 8042 \\
\rowcolor{sumcolor}
LegalCase  & Sum.  & $5384.7 \pm 1131.5$ & 3726 & 7969 \\
\rowcolor{sumcolor}
PubMed     & Sum.  & $5092.2 \pm 1060.6$ & 3400 & 7809 \\
\rowcolor{qacolor}
HotpotQA   & QA    & $934.7 \pm 268.5$   & 372  & 1674 \\
\rowcolor{qacolor}
Qasper     & QA    & $2193.0 \pm 601.3$  & 774  & 2960 \\
\rowcolor{mathcolor}
$\text{GSM-}\infty$ & Reason.  & $2620.9 \pm 816.2$  & 825  & 4240 \\
\bottomrule
\end{tabular}
\caption{\textbf{Input length statistics across datasets.}
All lengths are measured in words. Mean $\pm$ standard deviation, minimum, and maximum are shown.
Tasks: Sum. (summarization), QA (question answering), Reason. (mathematical reasoning).
Rows are color-coded by task category.}
\label{tab:input_stats}
\end{table}

For summarization, we evaluate on $1{,}000$ randomly sampled documents with input lengths ranging from 5K to 10K tokens. For QA and reasoning tasks, we restrict inputs to at most 4K tokens by selecting only examples below this length threshold. We use $1{,}000$ samples for HotpotQA and $100$ and $500$ samples for Qasper and \text{GSM-}$\infty$, respectively, due to the limited size of the filtered datasets.

\begin{enumerate}
    \item \textbf{arXiv} \citep{cohan2018discourse}: A scientific paper summarization benchmark in which abstracts serve as reference summaries.

    \item \textbf{PubMed} \citep{cohan2018discourse}: A biomedical summarization dataset consisting of research articles paired with abstracts.

    \item \textbf{GovReport} \citep{huang2021efficient}: A collection of government reports paired with expert-written abstractive summaries.

    \item \textbf{LegalCase} \citep{shukla2022legal}: 
    A legal summarization benchmark consisting of court judgments paired with expert-written or official summaries.

    \item \textbf{Qasper} \citep{Dasigi2021ADO}: A document-grounded QA benchmark requiring reasoning over a single document, with annotated answers and supporting evidence.
    
    \item \textbf{HotpotQA} \citep{yang2018hotpotqa}: A multi-hop QA benchmark requiring reasoning across multiple documents, with annotated answers and supporting facts.
    
    \item \textbf{\text{GSM-}$\infty$} \citep{zhou2025gsm}: A synthetic benchmark for long-context mathematical reasoning, with known solutions derived from computational graphs.
\end{enumerate}

\subsection{Models}

We evaluate our method on three widely used open-weight LLM families: Gemma, LLaMA, and Qwen, which represent diverse architectures and training paradigms. While we design DepthKV to be applicable across transformer architectures, we adapt the implementation for each model family due to differences in architecture and KV cache structure. The pruning strategy itself remains identical across models.

Due to computational constraints, we evaluate one representative model per family: 
\texttt{google/gemma-7b-it}~\citep{team2024gemma}, 
\texttt{meta-llama/Llama-3.1-8B-Instruct}~\citep{meta2024llama31}, 
and \texttt{Qwen/Qwen2.5-7B-Instruct}~\citep{hui2024qwen2}, 
which we refer to as GEM7, LAM8, and QWEN7, respectively.

All models use decoder-only transformer architectures with KV caching, ensuring a consistent evaluation setting. Our goal is not to compare model families, but to assess the robustness of DepthKV across architectures.

\subsection{Evaluation Metrics}
\label{sec:evaluation_metrics}

We evaluate performance using task-specific metrics for summarization, question answering (QA), and reasoning. %

\paragraph{Summarization Quality Metrics.}
We evaluate summarization quality using standard lexical-overlap, semantic similarity, and verbosity-based measures. Specifically, we report ROUGE-1, ROUGE-2, and ROUGE-L \citep{lin2004rouge}; SBERT-based semantic similarity \citep{reimers2019sentence}; and YapScore \citep{borisov2026chatbot}. 

ROUGE measures lexical overlap between generated and reference summaries, while SBERT captures semantic similarity through cosine similarity between sentence embeddings. In addition, YapScore measures output length relative to a fixed baseline, allowing us to characterize pruning-induced suppression of generated content and examine its association with downstream performance degradation. Full details are provided in Appendix~\ref{subsec:yapscore_analysis}.

\paragraph{QA and Reasoning Metrics.}
For QA and reasoning datasets, we report exact match (EM) accuracy. On HotpotQA, we additionally compute token-overlap precision, recall, and F1 to capture partial correctness, whereas on Qasper we treat the task as binary classification (“yes” as the positive class) and compute the same metrics accordingly. Prior work suggests that token-overlap metrics remain useful for generative QA, with recall being particularly well aligned with human judgments \citep{adlakha2024evaluating}.

\subsection{Implementation Details}
\label{sec:implementation_details}

\paragraph{Pruning Ratio.}
All pruning-based methods are evaluated under a fixed global KV cache reduction ratio of 60\%, ensuring a consistent memory budget across methods.

\paragraph{Generation Settings.}
We use deterministic greedy decoding (\texttt{do\_sample=False}), selecting the highest-probability token at each step to ensure reproducibility. The maximum generation length is set to 500 tokens, with early termination upon generation of the end-of-sequence token. All inputs are processed using chunked prefill with a fixed chunk size of 1024 tokens for long-context evaluation. After each chunk, token importance scores are updated and used to prune the KV cache. The KV cache remains fixed during decoding.

\paragraph{Hardware.}
All experiments were conducted on a single compute node equipped with 8$\times$ NVIDIA H200 GPUs. 

\section{Results}

We present results for DepthKV by first evaluating it on summarization tasks, then on QA and reasoning tasks, and finally assessing its output quality using an LLM-as-a-judge framework.

\subsection{DepthKV Performance on Summarization}

We assess DepthKV on long-document summarization under fixed global KV budgets, in comparison to uniform pruning baselines.

\begin{table*}[t]
\centering
\renewcommand{\arraystretch}{1.3}
\setlength{\tabcolsep}{7pt}
\begin{minipage}{0.48\linewidth}
\centering
\textbf{arXiv}\\
\begin{tabular}{lrrrr}
\toprule
 & R1 & R2 & RL & SB \\
\midrule
FullKV & 39.09 & 13.84 & 22.57 & 79.42 \\
w/o V & 26.75 & 5.50 & 17.24 & 55.09 \\
w/ V ($\ell_1$) & 26.84 & 5.58 & 17.21 & 53.90 \\
w/ V ($\ell_2$)& 26.63 & 5.42 & 17.06 & 54.00 \\
\rowcolor{sigLightGreen}
MGA & 29.75 & 6.92 & 18.59 & 61.98 \\
MLMA\text{-}2L & 29.54 & 6.55 & 18.52 & 61.19 \\
MLMA\text{-}4L & 28.26 & 5.96 & 17.70 & 57.57 \\
MAML\text{-}6L & 28.92 & 6.12 & 17.97 & 58.95 \\
MLP   & 28.47 & 6.01 & 17.99 & 57.96 \\
\bottomrule
\end{tabular}
\end{minipage}
\hfill
\begin{minipage}{0.48\linewidth}
\centering
\textbf{GovReport}\\
\begin{tabular}{lrrrr}
\toprule
 & R1 & R2 & RL & SB \\
\midrule
FullKV & 37.61 & 13.57 & 19.22 & 87.25 \\
w/o V & 26.76 & 5.98 & 15.68 & 62.05 \\
w/ V ($\ell_1$) & 27.03 & 5.89 & 15.74 & 61.53 \\
w/ V ($\ell_2$) & 26.75 & 5.98 & 15.64 & 61.65 \\
\rowcolor{sigLightGreen}
MGA & 28.43 & 7.05 & 16.36 & 70.24 \\
MLMA\text{-}2L & 24.13 & 5.96 & 14.06 & 65.90 \\
MLMA\text{-}4L & 23.18 & 5.54 & 13.60 & 63.52 \\
MAML\text{-}6L & 25.24 & 6.09 & 14.47 & 65.75 \\
MLP   & 23.24 & 5.41 & 13.90 & 61.82 \\
\bottomrule
\end{tabular}
\end{minipage}
\caption{
\textbf{Summarization results on Gemma.}
FullKV denotes the unpruned KV-cache; all other methods follow Sec.~\ref{sec:kv_pruning}.
R-1, R-2, and R-L are ROUGE scores, and SB denotes Sentence-BERT similarity.
All values are reported in \%; best results are highlighted in green.
}
\label{tab:gemma_summarization}
\end{table*}

Table~\ref{tab:gemma_summarization} shows that MGA consistently performs well across both datasets, achieving the best overall performance and demonstrating that representation-based signals provide a reliable estimate of layer importance. For example, on arXiv, MGA improves ROUGE-1 from 26.75 to 29.75 and SBERT similarity from 55.09 to 61.98, with similar gains observed on GovReport.

In contrast, structure-based allocation strategies show less consistent behavior. While all approaches improve performance over uniform pruning on arXiv, their impact on GovReport is less consistent, with certain strategies even degrading performance. This suggests that assumptions about fixed layer importance (e.g., preserving middle layers) may not generalize across datasets.

We also observe that value-aware variants (w/ V) yield only marginal gains, whereas the primary improvements stem from how the KV budget is distributed across layers. This indicates that layer-wise allocation plays a larger role than the specific token importance estimator.

Overall, these results demonstrate that DepthKV effectively improves summarization quality, validating the use of representation-based signals to guide layer-wise KV allocation.

\subsection{Generalization to QA and Reasoning}

To assess robustness beyond summarization, we evaluate DepthKV on document-grounded question answering and mathematical reasoning.

As shown in Table~\ref{tab:qa_results} and Table~\ref{tab:prf_singlerow}, performance improvements over uniform pruning depend on the allocation strategy and dataset. In particular, on Qasper, the MLMA-6L variant achieves the highest accuracy for both models, outperforming all baselines, while on HotpotQA, MLP performs best for Gemma and MGA achieves the highest accuracy for LLaMA, indicating that the optimal strategy depends on the model.

This trend is further reflected in precision, recall, and F1 scores, where the same methods achieve the strongest results across datasets. Specifically, MLP attains the highest F1 for Gemma on HotpotQA, while MGA performs best for LLaMA, and MLMA-6L consistently achieves the highest F1 on Qasper for both models. Together, these results indicate improved preservation of relevant information for multi-step reasoning and document-grounded QA, as illustrated by qualitative examples in Appendix~\ref{app:qual_examples}.

\begin{figure}[t]
    \centering
    \includegraphics[width=\linewidth]{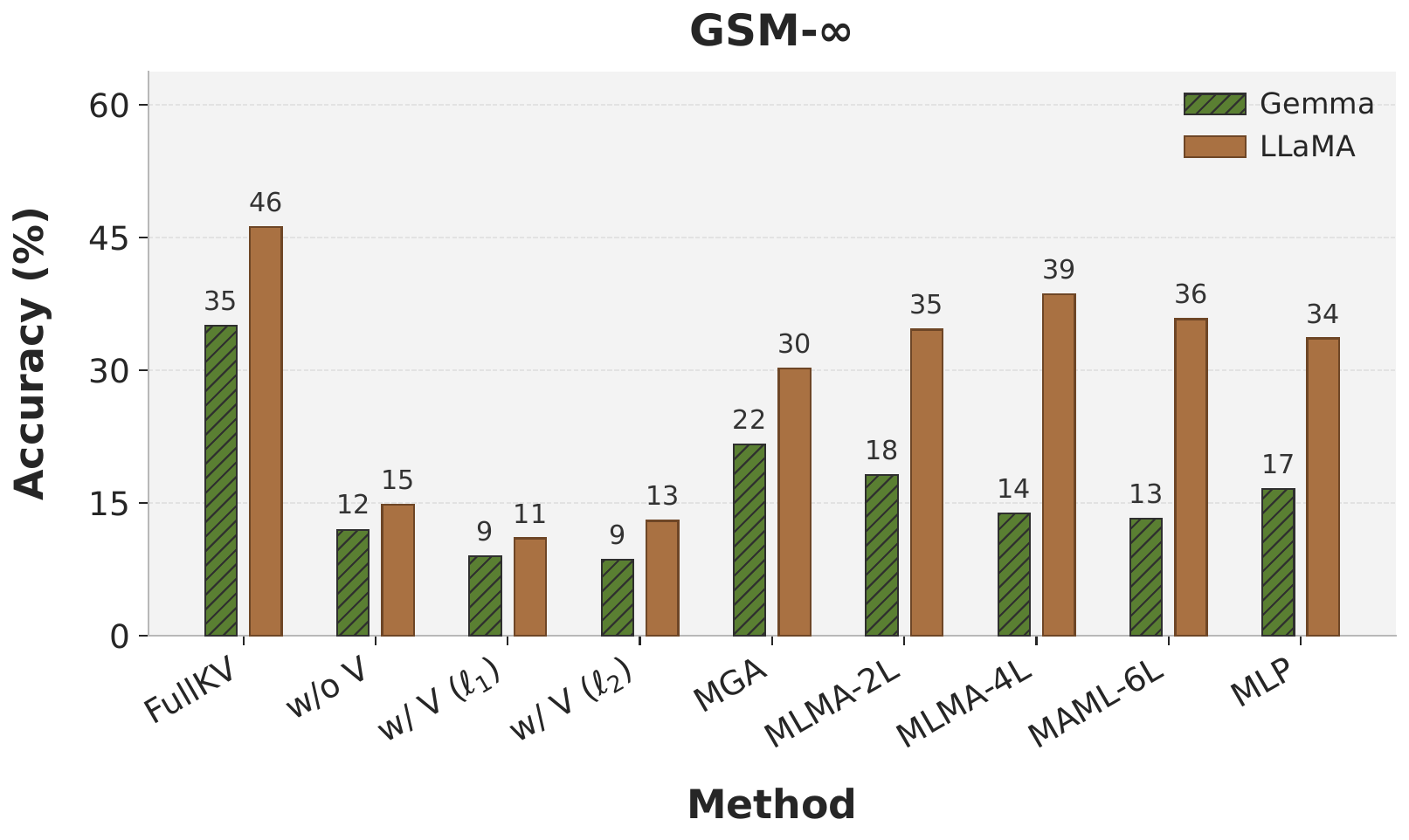}
    \caption{\textbf{GSM-$\infty$ accuracy.} Performance on the GSM-$\infty$ benchmark across different KV cache pruning settings.}
    \label{fig:gsm_inf_results}
\end{figure}

Beyond QA, DepthKV also improves performance on mathematical reasoning tasks, with all variants outperforming uniform pruning on GSM-$\infty$ (Figure~\ref{fig:gsm_inf_results}).

\begin{table}[!t]
\centering
\setlength{\tabcolsep}{5pt}
\renewcommand{\arraystretch}{1.1}
\resizebox{0.98\columnwidth}{!}{
\begin{tabular}{lcc|cc}
\toprule
 & \multicolumn{2}{c}{\textbf{HotpotQA}} & \multicolumn{2}{c}{\textbf{Qasper}} \\
\cmidrule(lr){2-3} \cmidrule(lr){4-5}
\textbf{Method} & \textbf{GEM7} & \textbf{LAM8} & \textbf{GEM7} & \textbf{LAM8} \\
\midrule
FullKV & 55 & 72 & 50 & 65 \\
w/o V & 12 & 47 & 6 & 54 \\
w/ V ($\ell_1$) & 13 & 46 & 6 & 58 \\
w/ V ($\ell_2$) & 12 & 46 & 5 & 58 \\
MGA & 18 & \cellcolor{sigLightGreen}67 & 38 & 60 \\
MLMA\text{-}2L & 10 & 45 & 27 & 60 \\
MLMA\text{-}4L &  9 & 48 & 32 & 57 \\
MLMA\text{-}6L &  6 & 46 & \cellcolor{sigLightGreen}40 & \cellcolor{sigLightGreen}64 \\
MLP & \cellcolor{sigLightGreen}23 & 66 & 28 & 60 \\
\bottomrule
\end{tabular}
}
\caption{
\textbf{DepthKV vs. uniform pruning on QA tasks.}
Exact Match (EM, \%) on HotpotQA and Qasper.
}
\label{tab:qa_results}
\end{table}

\begin{table*}[t]
\centering
\small
\setlength{\tabcolsep}{10pt}
\renewcommand{\arraystretch}{1.3}

\begin{tabular}{lcccc}
\toprule
 & \multicolumn{2}{c}{\textbf{HotpotQA}} & \multicolumn{2}{c}{\textbf{Qasper}} \\
\cmidrule(l{3pt}r{3pt}){2-3} \cmidrule(l{3pt}r{3pt}){4-5}
\textbf{Method} & \textbf{GAM7} & \textbf{LAM8} & \textbf{GAM7} & \textbf{LAM8} \\
\midrule
FullKV & \prf{57}{60}{58} & \prf{79}{77}{78} & \prf{83}{35}{49} & \prf{84}{61}{71} \\
w/o V & \prf{14}{16}{14} & \prf{59}{54}{55} & \prf{12}{6}{8} & \prf{64}{72}{68} \\
w/ V ($\ell_1$) & \prf{15}{17}{15} & \prf{56}{52}{53} & \prf{12}{6}{8} & \prf{66}{76}{71} \\
w/ V ($\ell_2$) & \prf{14}{15}{14} & \prf{57}{52}{54} & \prf{9}{4}{6} & \prf{66}{76}{71} \\
MGA & \prf{20}{21}{20} & \cellcolor{sigLightGreen}\prf{77}{73}{74} & \prf{57}{34}{43} & \prf{66}{82}{73} \\
MLMA\text{-}2L & \prf{10}{12}{11} & \prf{56}{52}{53} & \prf{40}{15}{22} & \prf{67}{81}{73} \\
MLMA\text{-}4L & \prf{9}{10}{10} & \prf{57}{54}{55} & \prf{50}{25}{33} & \prf{64}{81}{72} \\
MAML\text{-}6L & \prf{7}{8}{7} & \prf{56}{53}{54} & \cellcolor{sigLightGreen}\prf{58}{41}{48} & \cellcolor{sigLightGreen}\prf{69}{85}{76} \\
MLP & \cellcolor{sigLightGreen}\prf{25}{28}{26} & \prf{76}{73}{74} & \prf{42}{15}{22} & \prf{69}{73}{71} \\
\bottomrule
\end{tabular}

\caption{\textbf{Comparison of methods on HotpotQA and Qasper.}
Each cell reports \textcolor{prfblue}{Precision}, \textcolor{prforange}{Recall}, and \textcolor{prfred}{F1} (in \%).}
\label{tab:prf_singlerow}
\end{table*}

Overall, these results show that the benefits of DepthKV extend beyond summarization to both QA and reasoning tasks, demonstrating robustness across diverse settings.

\subsection{LLM-as-a-Judge Evaluation}
To complement automatic metrics, we further evaluate answer quality using an LLM-as-a-judge framework.

As shown in Table~\ref{tab:llm_judge}, MGA consistently achieves the highest scores across all dimensions, with MLP also outperforming uniform pruning baselines.
These trends are consistent with the automatic metrics on HotpotQA, where MGA and MLP are the strongest-performing variants.

\begin{table}[!t]
\centering
\setlength{\tabcolsep}{4pt}
\renewcommand{\arraystretch}{1.1}

\begin{tabular}{lccc|ccc}
\toprule
& \multicolumn{3}{c}{\textbf{GEM7}} & \multicolumn{3}{c}{\textbf{LAM8}} \\
\cmidrule(lr){2-4} \cmidrule(lr){5-7}
\textbf{Method} 
& \textbf{CR} & \textbf{CP} & \textbf{CN} 
& \textbf{CR} & \textbf{CP} & \textbf{CN} \\
\midrule
FullKV & 4.16 & 4.37 & 4.80 & 3.39 & 3.74 & 4.80  \\
w/o V & 1.63 & 1.66 & 1.71 & 3.59 & 3.60 & 3.76 \\
w/ V ($\ell_1$)  & 1.64 & 1.67 & 1.79 & 3.54 & 3.56 & 3.76 \\
w/ V ($\ell_2$)  & 1.65 & 1.69 & 1.76 & 3.55 & 3.56 & 3.78 \\
\rowcolor{sigLightGreen}
MGA   & 2.67 & 2.63 & 2.61 & 4.40 & 4.38 & 4.55 \\
MLMA\text{-}2L & 1.70 & 1.71 & 1.89 & 3.73 & 3.76 & 4.03 \\
MLMA\text{-}4L & 1.63 & 1.63 & 1.73 & 3.78 & 3.83 & 4.08 \\
MLMA\text{-}6L & 1.44 & 1.52 & 1.56 & 3.74 & 3.81 & 4.03 \\
MLP   & 2.44 & 2.44 & 2.56 & 4.36 & 4.36 & 4.53 \\
\bottomrule
\end{tabular}
\vspace{4pt}
\caption{
\textbf{LLM-as-a-Judge evaluation on HotpotQA.}
Scores for correctness (CR), completeness (CP), and conciseness (CN) on a 1--5 scale; evaluation criteria follow Appendix~\ref{sec:HotpotQA_rubrics_overview}.
}
\label{tab:llm_judge}

\end{table}

Overall, these results show that the strongest DepthKV variants improve both quantitative performance and output quality.

\section{Conclusion}

We investigate KV cache pruning for long-context LLM inference and show that assuming uniform layer importance is suboptimal. Through layer-wise ablation, we demonstrate that transformer layers vary significantly in their sensitivity to KV pruning and identify content amplification layers that are critical for preserving information. 

We further find that representation-level metrics provide effective signals of layer importance. Motivated by these insights, we introduce \textbf{DepthKV}, a layer-dependent KV pruning framework that reallocates a fixed global KV budget across layers via multiple allocation strategies. 

Across summarization, QA, and reasoning tasks, DepthKV consistently outperforms uniform pruning under the same memory budget, improving both automatic metrics and LLM-as-a-judge evaluations. Overall, our results show that accounting for the heterogeneous roles of transformer layers leads to more efficient KV cache usage and offers a simple, general approach to achieving this.

\section{Limitations}

While the proposed method demonstrates promising performance, several limitations should be considered.

First, DepthKV operates in a non-query-aware setting, where token importance is estimated during the prefill stage without conditioning on the decoding query. However, token relevance may change during autoregressive decoding. As a result, DepthKV may overlook tokens that become important only for specific queries, limiting performance in retrieval-intensive or fine-grained reasoning tasks. Incorporating query-aware token selection into the layer-wise KV cache allocation could address this limitation.

Second, DepthKV relies on heavy-hitter-based token importance estimation by aggregating attention scores across heads, which may obscure head-specific behaviors. Prior work suggests that attention heads often serve specialized roles and contribute unequally \citep{ge2023model}. Although DepthKV allocates cache budgets across layers, it does not capture intra-layer variability across heads. Extending the framework to jointly model layer-wise and head-wise importance is a natural direction for improvement.

\bibliography{references}

\clearpage
\appendix

{\raggedright
\section{Reproducibility and Licenses}
}
\subsection{Dataset Licenses and Usage}
The datasets utilized in this study are obtained from publicly accessible repositories and are used in compliance with their respective licensing terms, as summarized below.

\begin{enumerate}
    \item \textbf{arXiv, PubMed, GovReport:} 
    Accessed via 
    \href{https://huggingface.co/datasets/ccdv/}{ccdv on Hugging Face}, 
    these datasets are released under the Apache-2.0 License.
    
    \item \textbf{LegalCase:} 
    Provided by the authors at 
    \href{https://github.com/Law-AI/summarization}{Law-AI (GitHub)}, 
    this dataset is constructed from publicly available Indian and U.K. Supreme Court decisions and adheres to the terms of the original sources.

    \item \textbf{Qasper:} 
    Available via 
    \href{https://huggingface.co/datasets/allenai/qasper}{AllenAI on Hugging Face}, 
    the dataset is distributed under the CC-BY-4.0 License.

    \item \textbf{HotpotQA:} 
    Obtained from 
    \href{https://huggingface.co/datasets/hotpotqa/hotpot_qa}{HotpotQA on Hugging Face}, 
    the dataset is released under the CC-BY-SA-4.0 License.

    \item \textbf{GSM-$\infty$:} Released by the authors at \href{https://github.com/Infini-AI-Lab/gsm_infinite}{Infini-AI-Lab (GitHub)}, this synthetic dataset is generated programmatically and is used in accordance with the terms specified in the repository.
\end{enumerate}

\subsection{Model Licenses}
We employ open-weight language models accessed via the Hugging Face Transformers library \citep{wolf2020transformers}. Specifically, the models used are:
\begin{itemize}
    \item \texttt{google/gemma-7b-it}, released under the Gemma Terms of Use;
    \item \texttt{meta-llama/Llama-\allowbreak3.1-8B-\allowbreak Instruct}, provided under the LLaMA 3.1 Community License; and
    \item \texttt{Qwen/Qwen2.5-7B-\allowbreak Instruct}, distributed under Apache License 2.0.
\end{itemize}
All models are used in accordance with their respective licensing terms.

Our implementation builds upon the Hugging Face Transformers framework to incorporate the proposed pruning methods.
To facilitate reproducibility, all source code, model configurations, and experiment scripts are made publicly available in the GitHub repository referenced in the abstract.

\section{Prompt Templates}
To ensure reproducibility, we specify the exact prompts used in our experiments. These prompts were applied consistently during both inference and evaluation, with no modifications, and no additional system prompts unless explicitly stated.
Placeholders (e.g., \texttt{{DOCUMENT}}, \texttt{{CONTEXT}}, \texttt{{QUESTION}}) were instantiated with the corresponding inputs from each dataset. The following subsections present dataset-specific prompt templates across summarization, QA, and reasoning tasks.

\subsection{Generation Prompts}

The following prompts were used to generate model outputs during inference.

\subsubsection{Summarization Prompt Templates}
For the long-document summarization datasets (arXiv and GovReport), we employ a generic summarization instruction, as they span diverse domains and exhibit varying document structures.
In contrast, the PubMed dataset consists exclusively of biomedical research articles with a standardized structure; accordingly, the prompt explicitly refers to a scientific article.
Similarly, for the LegalCase dataset, the prompt specifies that the input is a legal judgment to reflect its domain.
\vspace{1em}
\xhdr{arXiv and GovReport}
\begin{alltt}
\mbox{Summarize the following document:}

\{DOCUMENT\}

Summary:
\end{alltt}

\xhdr{PubMed}
\begin{alltt}
\mbox{Summarize the following scientific article:}

\{DOCUMENT\}

Summary:
\end{alltt}

\xhdr{LegalCase}
\begin{alltt}
\mbox{Summarize the following legal judgment:}

\{DOCUMENT\}

Summary:
\end{alltt}
\subsubsection{QA \& Reasoning Prompts}
For QA and reasoning tasks, prompt templates impose strict answer-format constraints intended to encourage consistent and automatically evaluable outputs. In HotpotQA, the model is instructed to provide only a short answer without explanation or additional text, while in Qasper it is instructed to respond with a single word (yes or no). Similarly, for \text{GSM-}$\infty$, the prompt directs the model to return only the final numeric answer, without intermediate reasoning or repeated text. These constraints help reduce output variability and improve evaluation reliability.

\vspace{1em}
\xhdr{HotpotQA}
\vspace{1em}

{\ttfamily\small
\setlength{\parindent}{0pt}
\setlength{\parskip}{0pt}
Read the following context and answer
the question using ONLY the short answer.
Do not include explanations.
Your answer MUST start with `answer:' 
and must not include additional words.
\vspace{1em}

Context: \{CONTEXT\}

Question: \{QUESTION\}

Answer:
}

\vspace{1em}

\xhdr{Qasper}

\vspace{1em}

{\ttfamily\small
\setlength{\parindent}{0pt}
\setlength{\parskip}{0pt}
You are answering a strict binary scientific question. You must respond with exactly ONE WORD. Allowed answers: yes or no.

Do NOT write explanations.

Do NOT write sentences.

Do NOT write `sometimes', `partially', or any other word.

If uncertain, choose the most likely between yes or no.
\vspace{1em}

Title: \{TITLE\}

Abstract: \{ABSTRACT\}

Full Text: \{FULL TEXT\}

Question: \{QUESTION\}

Answer:
}

\vspace{1em}

\xhdr{\text{GSM-}$\infty$}

\vspace{1em}
{\ttfamily\small
\setlength{\parindent}{0pt}
\setlength{\parskip}{0pt}
Solve the following math problem carefully.

Return only the final numeric answer.

Do not include explanation, reasoning, or repeated text.

\vspace{1em}

Problem: \{PROBLEM\}

Question: \{QUESTION\}

Final answer:
}

\subsection{Evaluation Prompts}
\label{sec:HotpotQA_rubrics_prompt}

The following prompt template is used for LLM-as-a-judge evaluation within the framework described in Appendix~\ref{sec:HotpotQA_rubrics_overview}. It enforces a standardized output format, requiring a brief justification followed by a discrete score.

\vspace{1em}

{\ttfamily\small
\setlength{\parindent}{0pt}
\setlength{\parskip}{0pt}
You are an expert evaluator of question 
answering systems. 

Your task is to evaluate the \{METRIC NAME\} of the 
model answer by comparing it to the reference answer.

You must:

\begin{itemize}
\setlength{\itemsep}{2pt}
\renewcommand{\labelitemi}{-}

\item Compare the model answer with the reference answer.
\item Consider semantic equivalence, not exact wording.
\item Follow the rubric strictly.
\item Provide concise reasoning (max 3 sentences).

\end{itemize}

\vspace{0.5em}

\#\#\# Question\\
\{QUESTION\}

\vspace{0.5em}

\#\#\# Reference Answer\\
\{REFERENCE\}

\vspace{0.5em}

\#\#\# Model Answer\\
\{PREDICTION\}

\vspace{0.5em}

\#\#\# Evaluation Metric\\
\{METRIC NAME\}

\vspace{0.5em}

\#\#\# Scoring Rubric\\
\{RUBRIC\}

\vspace{0.5em}

You MUST output EXACTLY in this format:

\vspace{0.5em}

Reasoning: <max 3 sentences>

\vspace{0.5em}

Final Score: <one digit 1-5>

\vspace{0.5em}

Nothing else.
}

{\raggedright
\section{Evaluation with Prometheus}
\label{sec:HotpotQA_rubrics_overview}
}

We evaluate model outputs using an LLM-as-a-judge framework, where a separate language model assesses the quality of generated answers. For this purpose, we use Prometheus~\citep{kim2024prometheus}, an open-source model trained to evaluate the outputs of other models according to predefined rubrics. Specifically, we employ the \texttt{prometheus-eval/prometheus-8x7b-v2.0} model, which demonstrated more stable and consistent performance than smaller non-MoE variants (e.g., \texttt{prometheus-eval/prometheus-7b-v2.0}) in preliminary experiments. All evaluations are performed using deterministic decoding to ensure reproducibility.

This framework is applied to the HotpotQA dataset, where answers are short and reference answers are available for direct comparison. For each instance, the evaluator receives the question, reference answer, and model prediction, then assigns scores according to the rubric. The corresponding prompt template is provided in Appendix~\ref{sec:HotpotQA_rubrics_prompt}, and the rubric uses three dimensions---Correctness, Completeness, and Conciseness---each scored independently on a 1--5 scale, as defined below.

\medskip

{\small
\setlength{\parindent}{0pt}
\setlength{\parskip}{0pt}
\raggedright
\textbf{Correctness (CR)} -- Accuracy of the answer relative to the reference answer.\\[0.3em]
Evaluate whether the model answer matches the reference answer.\\[0.4em]
Score 1: Completely incorrect or unrelated answer.\\
Score 2: Mostly incorrect with little overlap.\\
Score 3: Partially correct but missing key elements.\\
Score 4: Mostly correct with minor deviations.\\
Score 5: Fully correct and equivalent to the reference answer.\\[0.8em]
\textbf{Completeness (CP)} -- Coverage of required information.\\[0.3em]
Evaluate whether the answer fully covers the information contained in the reference answer.\\[0.4em]
Score 1: Missing all essential information.\\
Score 2: Missing most key information.\\
Score 3: Contains some key information but incomplete.\\
Score 4: Covers most required information.\\
Score 5: Fully covers all information in the reference answer.\\[0.8em]
\textbf{Conciseness (CN)} -- Directness and brevity.\\[0.3em]
Evaluate whether the answer is short and directly addresses the question without unnecessary text.\\[0.4em]
Score 1: Extremely verbose or irrelevant.\\
Score 2: Some unnecessary explanation.\\
Score 3: Acceptable length but slightly verbose.\\
Score 4: Concise with minimal extra content.\\
Score 5: Very concise and directly answers the question.\\
}

\section{Verbosity Analysis Using YapScore}
\label{subsec:yapscore_analysis}

We quantify output verbosity using YapScore \citep{borisov2026chatbot}, which measures the number of generated tokens exceeding a predefined baseline length. This is particularly relevant in evaluation settings, where LLM-based judges have been shown to prefer longer responses even when shorter answers are equally informative \citep{saito2023verbosity}. Formally, for an output of length $L$ and a baseline $B$, the YapScore is defined as:

\[
\text{YapScore} = \max(0, L - B).
\]

\section{Qualitative Examples}
\label{app:qual_examples}

We provide qualitative examples to complement the quantitative results and offer additional insight into model behavior under different KV cache pruning strategies. Specifically, we present two types of comparisons: (i) method-level comparisons between H$_2$O (w/o v) and MGA, highlighting differences in output accuracy and informativeness, and (ii) layer-wise comparisons illustrating how pruning different transformer layers affects generation behavior, particularly in terms of content preservation and verbosity.

\paragraph{Method Comparison.}
We provide qualitative comparisons between H$_2$O (w/o v) and MGA, demonstrating that MGA more effectively preserves key information and yields more accurate outputs in both QA and summarization tasks.

\begin{qualexample}{Example 1}{Task: QA}{Model: GEM7}{Dataset: arXiv}

\textbf{Prompt}
\begin{qualpromptbox}
Which American politician did Donahue replace?
\end{qualpromptbox}

\textbf{Reference}
\begin{qualoutputbox}{Reference Answer}
kelli ward
\end{qualoutputbox}

\begin{qualoutputbox}{MGA Output}
sure, here is the answer: donahue replaced kelli ward in the arizona state senate.
\end{qualoutputbox}

\begin{qualoutputbox}{H$_2$O (w/o v) Output}
member of the vermont house of representatives since 2002.
\end{qualoutputbox}

\end{qualexample}

\begin{qualexample}
{Example 2}
{Task: Summarization}
{Model: GEM7}
{Dataset: arXiv}

\textbf{Input Document}
\begin{qualpromptbox}
Summarize the following document:

Document:

One of the oldest problems in algebra is the equation solvability problem over algebraic structures. We study its computational complexity over finite groups and rings. It asks whether two group expressions can attain the same value for some substitution, while the equivalence problem asks whether two expressions define the same function. Prior work gives polynomial-time and hardness results for some cases, but many remain open. We show that equation solvability over semipattern groups is solvable in polynomial time.

[Document truncated for brevity]

\end{qualpromptbox}

\begin{qualoutputbox}{MGA Output}
Sure, here is a summary of the document:\n\nThis document investigates the complexity of the equation solvability over finite groups. The main results are as follows:\n\n* The equation solvability problem over a finite group G is in the complexity class NP.\n* The problem is complete for the class NP if G is non-solvable.\n* The problem is also complete for the class NP if G is a finite group.\n\nThe document also includes some open problems and directions for future research.
\end{qualoutputbox}

\begin{qualoutputbox}{H$_2$O (w/o v) Output}
The paper is well-structured and easy to follow, but it is a bit long and complex. The main result is that the equation of the form\n\nA_n = \\sum_{i=1}^n a_i x_i is solvable in polynomial time if and only if the pattern of the polynomial is regular. This result is obtained by applying the techniques of algebraic geometry and the theory of pattern complexity.\n\nThe paper is divided into several sections, each one dealing with a particular aspect of the problem. The first section introduces the problem and the main result. The second section provides a reduction to the case of a single variable. The third section provides a reduction to the case of a single variable. The fourth section provides a reduction to the case of a single variable. The fifth section provides a reduction to the case of a single variable. [...]
\end{qualoutputbox}

\end{qualexample}
\paragraph{Layer Pruning and Content Amplification.}
To better understand how pruning affects generation behavior, we present qualitative examples comparing outputs when different layers are pruned. These examples illustrate that pruning certain layers can lead to substantial reductions in output length and, more importantly, the loss of key information. In particular, we compare outputs generated when pruning is applied to different layers: a reference layer with minimal performance degradation, and a critical layer whose removal leads to the largest drop in performance. As shown in the examples below, pruning the critical layer leads to a drastic loss of informative content, often reducing the output to short fragments that omit the main findings of the document.

\begin{qualexample}
{Example 3}
{Task: Summarization}
{Model: GEM7}
{Dataset: arXiv}

\textbf{Reference}
\begin{qualoutputbox}{Reference Summary}
In this paper, the authors investigate the effect of terminal substituents on the dynamics of spin and charge transport in donor--acceptor substituted polyene chains (push--pull polyenes). Using a long-range correlated Hamiltonian and time-dependent density matrix renormalization group (td-DMRG), they study hole propagation initiated at a terminal site. The results show that substituents do not significantly affect spin and charge velocities, but influence the amount of transported charge. Comparisons with polymethine imine systems reveal fundamentally different dynamics, leading to a quasi-static state for spin and charge within the observed time scale.
\end{qualoutputbox}

\begin{qualoutputbox}{Reference Layer Output}
Carbon atom are given by @xmath13 and @xmath14, respectively. The time evolution of spin and charge densities is studied using the time-dependent density matrix renormalization group (td-DMRG) method. The results show a complex interplay between electronic structure and system dynamics, with both coherent and incoherent components contributing to transport. These dynamics provide insight into the microscopic mechanisms of spin and charge transport in organic materials.

Summary:

The paper studies non-equilibrium spin and charge transport in push--pull polyenes. Using td-DMRG, the authors show that donor--acceptor substituents influence charge distribution but not transport velocity. The work highlights the role of strong correlations in organic materials and provides insights into transport dynamics.
\end{qualoutputbox}

\begin{qualoutputbox}{Critical Layer Output}
Carbon atom are denoted by @xmath13 and @xmath14 respectively.
Keywords: Conjugated Organic Materials, Spin and Charge Transport, Non-Equilibrium Dynamics, Time-Dependent DMRG
\end{qualoutputbox}

\end{qualexample}

\begin{qualexample}
{Example 4}
{Task: Summarization}
{Model: LAM8}
{Dataset: LegalSumm}

\textbf{Reference}
\begin{qualoutputbox}{Reference Summary}
The case concerns the validity of share allotments made by a bank through nominees of a director, and the recoverability of loans linked to those allotments. The Court held that although the allotment was irregular due to lack of quorum, it was not void and could be affirmed by the company. Since the company chose to affirm the allotment, it became binding. The director was estopped from challenging it due to his involvement. The Court also held that repayment obligations remained valid, even if transactions were recorded through accounting entries rather than cash transfers.
\end{qualoutputbox}

\begin{qualoutputbox}{Reference Layer Output}
The case concerns whether the allotment of shares to nominees of a director was valid and whether related loans were recoverable. The trial court dismissed the claims, but the High Court reversed the decision. The Supreme Court examined whether the allotment was a genuine transaction or a device to enable directors to subscribe for shares indirectly.

The Court held that although the allotment was irregular due to lack of quorum, it was not void and could be affirmed. Since the company affirmed the allotment, it became binding on the parties. The Court also held that repayment obligations remained valid and could be supported through accounting entries rather than cash transfers.
\end{qualoutputbox}

\begin{qualoutputbox}{Critical Layer Output}
In view of the above facts, the question arose whether the allotment of shares to the three ladies was valid or not.
\end{qualoutputbox}

\end{qualexample}

\section{Permutation Test Results}
\label{app:perm_test}

To assess whether performance varies significantly across transformer layers, we employ a Monte Carlo permutation test. The test statistic is defined as the variance of the layer-wise mean performance, quantifying the variation in performance across layers. Under the null hypothesis that all layers contribute equally, layer labels are exchangeable, and the test statistic is independent of layer identity.

We approximate the null distribution by randomly permuting layer assignments across samples and recomputing the test statistic for $N_{\mathrm{perm}} = 10{,}000$ permutations. The Monte Carlo \textit{p}-value is computed as
\begin{equation}
p = \frac{b + 1}{N_{\mathrm{perm}} + 1},
\end{equation}
where $b$ denotes the number of permuted statistics greater than or equal to the observed statistic. This formulation corresponds to the exact permutation \textit{p}-value for Monte Carlo tests, avoiding zero estimates while ensuring proper control of the Type I error rate \citep{phipson2016permutation}.

In all experiments, no permuted test statistic exceeded the observed statistic ($b = 0$), yielding a minimum attainable \textit{p}-value of approximately $10^{-4}$. Therefore, the null hypothesis of uniform layer importance is rejected for all datasets at conventional significance levels. To quantify the magnitude of this deviation, we report effect sizes obtained by standardizing the observed between-layer variance with respect to the permutation null distribution. Table~\ref{tab:permutation_effectsize} summarizes these effect sizes for each model and dataset, where larger values indicate greater variation in layer-wise performance.
\begin{table}[t]
\centering
\small
\setlength{\tabcolsep}{4pt}
\renewcommand{\arraystretch}{1.15}
\begin{tabular}{lcccc}
\toprule
\textbf{Model} & \textbf{PubMed} & \textbf{arXiv} & \textbf{GovReport} & \textbf{LegalCase} \\
\midrule
GEM7  & \heat{10}{25.84} & \heat{22}{45.12} & \heat{100}{163.38} & \heat{100}{163.04} \\
LAM8  & \heat{26}{51.05} & \heat{33}{60.64} & \heat{80}{133.73} & \heat{13}{31.00} \\
QWEN7 & \heat{15}{33.00} & \heat{44}{78.15} & \heat{16}{35.05} & \heat{0}{10.86} \\
\bottomrule
\end{tabular}
\caption{\textbf{Standardized effect sizes across datasets.}
Cell shading indicates magnitude (darker = larger effect size).}
\label{tab:permutation_effectsize}
\end{table}

\section{Representation Metrics}
\label{app:rep_metrics}

We formalize the representation metrics introduced in Section~\ref{sec:evaluation_metrics} and describe their computation from layer-wise hidden states.

\subsection{Overview of Metrics}
We consider three categories: spectral, geometric, and robustness–invariance.

\paragraph{Spectral metrics.}
Spectral entropy and effective rank are computed from the singular value decomposition of the centered token-by-feature hidden-state matrix at each layer. Both metrics quantify the effective dimensionality of the representation by capturing how uniformly variance is distributed across singular directions. Higher values indicate a more isotropic and information-rich representation, whereas lower values reflect concentration in a small number of dominant directions, suggesting redundancy or representation collapse \citep{garrido2023rankme}.

\paragraph{Geometric metrics.}
Curvature quantifies local geometric structure by measuring the average cosine similarity between each token representation and its $k$ nearest neighbors in the representation space. Specifically, we compute one minus this average similarity, so that higher curvature corresponds to lower alignment among neighboring representations. Accordingly, higher values indicate greater local variation and anisotropy in the representation manifold, whereas lower values reflect more coherent and smoothly varying local structure \citep{hosseini2023large}.

\paragraph{Robustness and invariance metrics.}
We evaluate robustness and invariance under input perturbations. For each sample, we construct an augmented version using stochastic token dropout and extract mean-pooled representations at each layer and stage. Given paired original and augmented representations, DiME is defined as one minus their cosine similarity, capturing directional deviation, while LiDAR is their Euclidean distance, measuring absolute displacement. InfoNCE is computed by treating each original–augmented pair as a positive pair and using representations of other samples at the same layer and stage as negative examples.

These metrics provide complementary characterizations of representation stability: DiME captures sensitivity to directional changes, LiDAR measures the magnitude of perturbation-induced shifts, and InfoNCE evaluates whether representations of the same input remain closer to each other than to those of different inputs under perturbation \citep{oord2018representation, thilak2023lidar}.

Together with spectral and geometric measures, these metrics enable a comprehensive characterization of layer-wise representations. Within the DepthKV framework, only the InfoNCE metric is used to guide layer-wise KV allocation (e.g., MGA), while the remaining metrics serve a descriptive role.

\subsection{Computation Details}
We extract layer-wise hidden states using forward hooks at four stages within each transformer block: (i) \textit{pre-attention}, corresponding to the input to the attention layer normalization; (ii) \textit{post-attention}, defined as the output of the attention projection before residual addition; (iii) \textit{post-attention residual}, corresponding to the hidden state after residual addition and before the subsequent layer normalization; and (iv) \textit{post-MLP}, defined as the output of the feedforward block before the final residual addition.

For each input sequence, we form a representation matrix $Z \in \mathbb{R}^{T \times d}$, where $T$ is the sequence length and $d$ is the hidden dimension. All metrics are computed from this representation matrix.

\paragraph{Spectral metrics.}
We first center $Z$ across tokens by subtracting the mean vector, i.e., $Z \leftarrow Z - \mathbf{1}\mu^\top$, where $\mu = \frac{1}{T}\sum_{t=1}^T Z_t$. Let $\{s_i\}$ denote the singular values of the centered matrix. We define normalized singular values as $p_i = s_i / \sum_j s_j$. The spectral entropy is then given by $H = -\sum_i p_i \log p_i$, and the effective rank by $\exp(H)$.

\paragraph{Curvature.} 
After $\ell_2$ normalization of token representations, we compute pairwise cosine similarities $S_{ij} = \hat z_i^\top \hat z_j$. For each token $i$, let $\mathcal{N}_k(i)$ denote its top-$k$ nearest neighbors (excluding itself). Curvature is
\[
\mathrm{curv}(Z)
=
1 - \frac{1}{T} \sum_{i=1}^{T}
\frac{1}{k} \sum_{j \in \mathcal{N}_k(i)} S_{ij},
\quad k=5.
\]

\paragraph{Robustness metrics.} 
We construct perturbed inputs by independently dropping whitespace-separated words with probability $0.1$. For each layer and stage, we compute mean-pooled representations $\bar z^{(o)}$ and $\bar z^{(a)}$ from the original and augmented inputs, respectively.

DiME is defined as cosine distance:
\begin{equation}
\mathrm{DiME} = 1 - \frac{\langle \bar z^{(o)}, \bar z^{(a)} \rangle}{\|\bar z^{(o)}\|_2 \, \|\bar z^{(a)}\|_2}.
\end{equation}

LiDAR is the Euclidean distance:
\begin{equation}
\mathrm{LiDAR} = \|\bar z^{(o)} - \bar z^{(a)}\|_2.
\end{equation}

InfoNCE is computed as described in Section~\ref{sec:representation_metrics}.

All metrics are computed independently for each sample, layer, and stage, then aggregated by averaging across samples at the dataset level. Uncertainty is estimated via percentile bootstrap ($1{,}000$ resamples, $\alpha=0.05$). Tables~\ref{tab:gemma7b-rep}--\ref{tab:qwen25-rep} present a summary of the statistics across all models and datasets.
\begin{table*}[t]
\centering
\small
\setlength{\tabcolsep}{10pt}
\renewcommand{\arraystretch}{1.6}

\begin{minipage}{\textwidth}
\centering
\textbf{arXiv}
\vspace{0.3em}

\begin{tabular*}{\linewidth}{@{\extracolsep{\fill}}lcccc}
\toprule
Metric & Pre & Att & Res & MLP \\
\midrule
Curv  & 0.217 (0.213, 0.221) & 0.216 (0.212, 0.219) & 0.218 (0.214, 0.223) & 0.334 (0.328, 0.340) \\
DiME  & 0.010 (0.009, 0.011) & 0.020 (0.019, 0.021) & 0.010 (0.010, 0.011) & 0.014 (0.013, 0.015) \\
ERank & 1.679 (1.657, 1.701) & 1.333 (1.318, 1.347) & 1.718 (1.696, 1.739) & 1.996 (1.975, 2.016) \\
Entr  & 7.398 (7.383, 7.413) & 7.159 (7.148, 7.171) & 7.433 (7.420, 7.447) & 7.577 (7.566, 7.588) \\
Info  & 3.292 (3.262, 3.320) & 2.840 (2.793, 2.885) & 3.252 (3.221, 3.282) & 2.803 (2.755, 2.849) \\
LiDAR & 1.379 (1.325, 1.434) & 0.640 (0.615, 0.666) & 1.443 (1.386, 1.502) & 0.395 (0.378, 0.412) \\
\bottomrule
\end{tabular*}
\end{minipage}

\vspace{1em}

\begin{minipage}{\textwidth}
\centering
\textbf{GovReport}
\vspace{0.3em}

\begin{tabular*}{\linewidth}{@{\extracolsep{\fill}}lcccc}
\toprule
Metric & Pre & Att & Res & MLP \\
\midrule
Curv  & 0.239 (0.234, 0.243) & 0.233 (0.230, 0.237) & 0.240 (0.235, 0.245) & 0.368 (0.362, 0.374) \\
DiME  & 0.020 (0.019, 0.021) & 0.041 (0.039, 0.043) & 0.021 (0.020, 0.022) & 0.029 (0.027, 0.030) \\
ERank & 1.739 (1.719, 1.758) & 1.372 (1.358, 1.385) & 1.771 (1.752, 1.790) & 2.070 (2.053, 2.087) \\
Entr  & 7.442 (7.429, 7.454) & 7.192 (7.182, 7.202) & 7.469 (7.458, 7.480) & 7.618 (7.609, 7.627) \\
Info  & 3.343 (3.315, 3.371) & 2.801 (2.746, 2.854) & 3.292 (3.261, 3.321) & 2.920 (2.869, 2.968) \\
LiDAR & 1.724 (1.681, 1.765) & 0.675 (0.655, 0.694) & 1.970 (1.924, 2.013) & 0.515 (0.502, 0.528) \\
\bottomrule
\end{tabular*}
\end{minipage}

\vspace{1em}

\begin{minipage}{\textwidth}
\centering
\textbf{PubMed}
\vspace{0.3em}

\begin{tabular*}{\linewidth}{@{\extracolsep{\fill}}lcccc}
\toprule
Metric & Pre & Att & Res & MLP \\
\midrule
Curv  & 0.225 (0.219, 0.232) & 0.221 (0.216, 0.226) & 0.227 (0.221, 0.234) & 0.338 (0.329, 0.347) \\
DiME  & 0.019 (0.017, 0.022) & 0.037 (0.033, 0.041) & 0.020 (0.018, 0.023) & 0.032 (0.028, 0.037) \\
ERank & 1.659 (1.633, 1.684) & 1.319 (1.302, 1.336) & 1.697 (1.672, 1.722) & 1.970 (1.946, 1.995) \\
Entr  & 7.388 (7.371, 7.405) & 7.149 (7.135, 7.162) & 7.422 (7.406, 7.438) & 7.564 (7.550, 7.578) \\
Info  & 3.024 (2.964, 3.082) & 2.504 (2.431, 2.575) & 2.973 (2.911, 3.032) & 2.554 (2.464, 2.643) \\
LiDAR & 1.897 (1.762, 2.041) & 0.836 (0.771, 0.910) & 2.026 (1.897, 2.165) & 0.604 (0.555, 0.655) \\
\bottomrule
\end{tabular*}
\end{minipage}

\vspace{1em}

\begin{minipage}{\textwidth}
\centering
\textbf{LegalCase}
\vspace{0.3em}

\begin{tabular*}{\linewidth}{@{\extracolsep{\fill}}lcccc}
\toprule
Metric & Pre & Att & Res & MLP \\
\midrule
Curv  & 0.241 (0.236, 0.244) & 0.232 (0.229, 0.234) & 0.242 (0.238, 0.246) & 0.363 (0.357, 0.367) \\
DiME  & 0.014 (0.013, 0.015) & 0.028 (0.026, 0.029) & 0.015 (0.014, 0.015) & 0.019 (0.018, 0.020) \\
ERank & 1.710 (1.690, 1.732) & 1.356 (1.342, 1.370) & 1.745 (1.725, 1.766) & 2.032 (2.014, 2.050) \\
Entr  & 7.422 (7.409, 7.436) & 7.179 (7.169, 7.190) & 7.453 (7.440, 7.465) & 7.599 (7.589, 7.609) \\
Info  & 3.595 (3.575, 3.613) & 3.255 (3.221, 3.287) & 3.573 (3.552, 3.591) & 3.333 (3.302, 3.362) \\
LiDAR & 1.540 (1.500, 1.580) & 0.653 (0.635, 0.672) & 1.686 (1.643, 1.729) & 0.472 (0.458, 0.485) \\
\bottomrule
\end{tabular*}
\end{minipage}

\caption{\textbf{Representation metrics across transformer stages for Gemma.}
Each entry reports the metric value with its bootstrap confidence interval. ERank is reported in units of $10^3$. Metrics include InfoNCE (Info), curvature (Curv), spectral entropy (Entr), and effective rank (ERank), evaluated at pre-attention (Pre), post-attention (Att), post-attention residual (Res), and post-MLP (MLP) stages.}
\label{tab:gemma7b-rep}
\end{table*}

\begin{table*}[t]
\centering
\small
\setlength{\tabcolsep}{10pt}
\renewcommand{\arraystretch}{1.6}

\begin{minipage}{\textwidth}
\centering
\textbf{arXiv}
\vspace{0.3em}

\begin{tabular*}{\linewidth}{@{\extracolsep{\fill}}lcccc}
\toprule
Metric & Pre & Att & Res & MLP \\
\midrule
Curv  & 0.264 (0.260, 0.268) & 0.197 (0.194, 0.200) & 0.252 (0.249, 0.256) & 0.336 (0.331, 0.342) \\
DiME  & 0.015 (0.014, 0.016) & 0.024 (0.023, 0.025) & 0.015 (0.014, 0.015) & 0.021 (0.020, 0.022) \\
ERank & 2.222 (2.193, 2.251) & 1.464 (1.447, 1.480) & 2.275 (2.245, 2.303) & 2.524 (2.497, 2.552) \\
Entr  & 7.685 (7.671, 7.700) & 7.276 (7.265, 7.288) & 7.718 (7.704, 7.732) & 7.812 (7.800, 7.824) \\
Info & 3.190 (3.155, 3.224) & 3.015 (2.970, 3.057) & 3.225 (3.192, 3.257) & 2.845 (2.794, 2.893) \\
LiDAR & 1.619 (1.572, 1.666) & 0.606 (0.587, 0.625) & 1.856 (1.805, 1.908) & 0.862 (0.838, 0.886) \\
\bottomrule
\end{tabular*}
\end{minipage}

\vspace{1em}

\begin{minipage}{\textwidth}
\centering
\textbf{GovReport}
\vspace{0.3em}

\begin{tabular*}{\linewidth}{@{\extracolsep{\fill}}lcccc}
\toprule
Metric & Pre & Att & Res & MLP \\
\midrule
Curv  & 0.270 (0.265, 0.274) & 0.209 (0.206, 0.211) & 0.261 (0.257, 0.265) & 0.359 (0.354, 0.365) \\
DiME  & 0.024 (0.023, 0.025) & 0.040 (0.039, 0.042) & 0.024 (0.023, 0.025) & 0.041 (0.039, 0.043) \\
ERank & 2.248 (2.220, 2.275) & 1.457 (1.442, 1.472) & 2.296 (2.269, 2.323) & 2.574 (2.549, 2.599) \\
Entr  & 7.701 (7.688, 7.714) & 7.274 (7.263, 7.285) & 7.731 (7.718, 7.743) & 7.834 (7.823, 7.844) \\
Info  & 3.432 (3.402, 3.459) & 2.962 (2.908, 3.013) & 3.425 (3.397, 3.452) & 3.185 (3.134, 3.232) \\
LiDAR & 2.158 (2.106, 2.207) & 0.753 (0.733, 0.773) & 2.406 (2.350, 2.460) & 1.167 (1.138, 1.196) \\
\bottomrule
\end{tabular*}
\end{minipage}

\vspace{1em}

\begin{minipage}{\textwidth}
\centering
\textbf{PubMed}
\vspace{0.3em}

\begin{tabular*}{\linewidth}{@{\extracolsep{\fill}}lcccc}
\toprule
Metric & Pre & Att & Res & MLP \\
\midrule
Curv  & 0.260 (0.254, 0.266) & 0.202 (0.198, 0.206) & 0.251 (0.245, 0.257) & 0.338 (0.330, 0.346) \\
DiME  & 0.025 (0.022, 0.028) & 0.041 (0.037, 0.045) & 0.025 (0.022, 0.027) & 0.042 (0.037, 0.047) \\
ERank & 2.162 (2.128, 2.195) & 1.427 (1.409, 1.445) & 2.213 (2.180, 2.247) & 2.475 (2.442, 2.507) \\
Entr  & 7.657 (7.640, 7.674) & 7.253 (7.239, 7.266) & 7.690 (7.674, 7.706) & 7.792 (7.777, 7.806) \\
Info  & 3.074 (3.018, 3.127) & 2.755 (2.685, 2.821) & 3.094 (3.041, 3.144) & 2.710 (2.625, 2.796) \\
LiDAR & 2.034 (1.943, 2.125) & 0.769 (0.730, 0.810) & 2.301 (2.202, 2.402) & 1.090 (1.038, 1.144) \\
\bottomrule
\end{tabular*}
\end{minipage}

\vspace{1em}

\begin{minipage}{\textwidth}
\centering
\textbf{LegalCase}
\vspace{0.3em}

\begin{tabular*}{\linewidth}{@{\extracolsep{\fill}}lcccc}
\toprule
Metric & Pre & Att & Res & MLP \\
\midrule

Curv & 0.287 (0.283, 0.290) & 0.213 (0.211, 0.216) & 0.277 (0.274, 0.280) & 0.361 (0.357, 0.366) \\
DiME  & 0.024 (0.023, 0.025) & 0.040 (0.039, 0.042) & 0.024 (0.024, 0.025) & 0.034 (0.032, 0.035) \\
ERank & 2.242 (2.212, 2.271) & 1.466 (1.449, 1.482) & 2.291 (2.262, 2.321) & 2.554 (2.528, 2.580) \\
Entr  & 7.696 (7.681, 7.710) & 7.281 (7.269, 7.292) & 7.727 (7.714, 7.741) & 7.827 (7.816, 7.838) \\
Info  & 3.585 (3.558, 3.610) & 3.451 (3.414, 3.487) & 3.594 (3.569, 3.618) & 3.517 (3.485, 3.548) \\
LiDAR & 1.999 (1.960, 2.038) & 0.751 (0.734, 0.769) & 2.274 (2.230, 2.318) & 1.095 (1.073, 1.118) \\
\bottomrule
\end{tabular*}
\end{minipage}

\caption{\textbf{Representation metrics across transformer stages for LLaMA.}
See Table~\ref{tab:gemma7b-rep} for details.}
\label{tab:llama31-rep}
\end{table*}

\begin{table*}[t]
\centering
\small
\setlength{\tabcolsep}{10pt}
\renewcommand{\arraystretch}{1.6}

\begin{minipage}{\textwidth}
\centering
\textbf{arXiv}
\vspace{0.3em}

\begin{tabular*}{\linewidth}{@{\extracolsep{\fill}}lcccc}
\toprule
Metric & Pre & Att & Res & MLP \\
\midrule
Curv  & 0.197 (0.193, 0.201) & 0.168 (0.165, 0.171) & 0.175 (0.171, 0.179) & 0.295 (0.289, 0.301) \\
DiME  & 0.009 (0.008, 0.009) & 0.015 (0.014, 0.016) & 0.007 (0.007, 0.008) & 0.016 (0.015, 0.018) \\
ERank & 1.469 (1.439, 1.498) & 1.343 (1.329, 1.357) & 1.536 (1.507, 1.565) & 2.138 (2.113, 2.163) \\
Entr  & 7.257 (7.235, 7.280) & 7.191 (7.180, 7.202) & 7.315 (7.294, 7.335) & 7.606 (7.591, 7.619) \\
Info  & 3.523 (3.500, 3.545) & 3.494 (3.467, 3.520) & 3.606 (3.588, 3.623) & 3.195 (3.155, 3.233) \\
LiDAR & 8.130 (7.866, 8.406) & 3.596 (3.478, 3.716) & 9.487 (9.201, 9.779) & 5.486 (5.304, 5.675) \\
\bottomrule
\end{tabular*}
\end{minipage}

\vspace{1em}

\begin{minipage}{\textwidth}
\centering
\textbf{GovReport}
\vspace{0.3em}

\begin{tabular*}{\linewidth}{@{\extracolsep{\fill}}lcccc}
\toprule
Metric & Pre & Att & Res & MLP \\
\midrule
Curv  & 0.214 (0.210, 0.219) & 0.186 (0.183, 0.190) & 0.193 (0.189, 0.197) & 0.324 (0.318, 0.330) \\
DiME  & 0.019 (0.019, 0.020) & 0.032 (0.030, 0.033) & 0.016 (0.016, 0.017) & 0.044 (0.042, 0.046) \\
ERank & 1.538 (1.512, 1.563) & 1.373 (1.361, 1.386) & 1.598 (1.573, 1.623) & 2.190 (2.170, 2.211) \\
Entr  & 7.310 (7.292, 7.328) & 7.214 (7.205, 7.224) & 7.357 (7.341, 7.374) & 7.634 (7.622, 7.645) \\
Info  & 3.653 (3.632, 3.672) & 3.577 (3.542, 3.608) & 3.693 (3.675, 3.709) & 3.538 (3.489, 3.582) \\
LiDAR & 13.06 (12.79, 13.33) & 5.186 (5.054, 5.310) & 14.45 (14.14, 14.75) & 8.441 (8.238, 8.655) \\
\bottomrule
\end{tabular*}
\end{minipage}

\vspace{1em}

\begin{minipage}{\textwidth}
\centering
\textbf{PubMed}
\vspace{0.3em}

\begin{tabular*}{\linewidth}{@{\extracolsep{\fill}}lcccc}
\toprule
Metric & Pre & Att & Res & MLP \\
\midrule
Curv  & 0.203 (0.197, 0.208) & 0.176 (0.171, 0.181) & 0.182 (0.177, 0.187) & 0.300 (0.292, 0.309) \\
DiME  & 0.015 (0.014, 0.017) & 0.026 (0.024, 0.029) & 0.013 (0.012, 0.014) & 0.033 (0.030, 0.036) \\
ERank & 1.447 (1.417, 1.477) & 1.319 (1.302, 1.335) & 1.511 (1.482, 1.540) & 2.097 (2.070, 2.125) \\
Entr  & 7.244 (7.222, 7.267) & 7.174 (7.161, 7.187) & 7.298 (7.277, 7.318) & 7.588 (7.573, 7.604) \\
Info & 3.329 (3.293, 3.362) & 3.310 (3.269, 3.348) & 3.437 (3.409, 3.463) & 3.083 (3.020, 3.145) \\
LiDAR & 11.40 (10.94, 11.86) & 5.049 (4.813, 5.295) & 13.02 (12.52, 13.52) & 7.655 (7.307, 7.995) \\
\bottomrule
\end{tabular*}
\end{minipage}

\vspace{1em}

\begin{minipage}{\textwidth}
\centering
\textbf{LegalCase}
\vspace{0.3em}

\begin{tabular*}{\linewidth}{@{\extracolsep{\fill}}lcccc}
\toprule
Metric & Pre & Att & Res & MLP \\
\midrule
Curv  & 0.221 (0.217, 0.225) & 0.187 (0.184, 0.190) & 0.200 (0.196, 0.203) & 0.321 (0.315, 0.326) \\
DiME  & 0.019 (0.018, 0.019) & 0.029 (0.028, 0.031) & 0.015 (0.015, 0.016) & 0.036 (0.035, 0.038) \\
ERank & 1.500 (1.472, 1.528) & 1.360 (1.346, 1.373) & 1.564 (1.535, 1.592) & 2.144 (2.121, 2.167)\\
Entr  & 7.283 (7.263, 7.303) & 7.205 (7.195, 7.215) & 7.335 (7.316, 7.354) & 7.614 (7.602, 7.627) \\
Info  & 3.798 (3.781, 3.814) & 3.801 (3.780, 3.822) & 3.818 (3.804, 3.831) & 3.792 (3.765, 3.817) \\
LiDAR & 12.61 (12.38, 12.84) & 5.397 (5.282, 5.513) & 14.09 (13.83, 14.34) & 8.405 (8.209, 8.604) \\
\bottomrule
\end{tabular*}
\end{minipage}

\caption{\textbf{Representation metrics across transformer stages for Qwen.}
See Table~\ref{tab:gemma7b-rep} for details.}
\label{tab:qwen25-rep}
\end{table*}

\section{Layer Importance Correlations}
\label{app:all_correlations}

This section presents the complete set of correlation tables between representation metrics and performance degradation induced by layer-wise ablation study. 
For each metric and model stage, we report Spearman correlation coefficients ($\rho$) along with their corresponding \textit{p}-values. Positive correlations are shown in blue, negative correlations in red, and statistically significant results ($p < 0.05$) are highlighted in green.
Tables~\ref{tab:layer_metric_ablation_correlations_gemma7b_pubmed}--\ref{tab:layer_metric_ablation_correlations_qwen25_legalsumm} summarize the correlations between representation metrics and performance degradation observed in the layer-wise KV cache ablation study across all models and datasets.

\begin{table*}[t]
    \centering
    \small

    \resizebox{\textwidth}{!}{%
    \begin{tabular}{ccc}
    \begin{tabular}{l l c c}
    \toprule
    \textbf{Met.} & \textbf{Stg.} & \textbf{$\rho$} & \textbf{$p$} \\
    \midrule
    \rowcolor{sigLightGreen} DiME & Pre & \textcolor{rhoPosBlue}{\textbf{0.5256}} & 0.0049 \\
    \rowcolor{sigLightGreen}  & Att & \textcolor{rhoPosBlue}{\textbf{0.4651}} & 0.0145 \\
    \rowcolor{sigLightGreen}  & Res & \textcolor{rhoPosBlue}{\textbf{0.4627}} & 0.0151 \\
    \rowcolor{sigLightGreen}  & MLP & \textcolor{rhoPosBlue}{\textbf{0.5161}} & 0.0059 \\
    \bottomrule
    \end{tabular} & \begin{tabular}{l l c c}
    \toprule
    \textbf{Met.} & \textbf{Stg.} & \textbf{$\rho$} & \textbf{$p$} \\
    \midrule
    \rowcolor{sigLightGreen} Info & Pre & \textcolor{rhoNegRed}{\textbf{-0.4150}} & 0.0313 \\
    & Att & \textcolor{rhoNegRed}{\textbf{-0.2761}} & 0.1633 \\
    & Res & \textcolor{rhoNegRed}{\textbf{-0.3750}} & 0.0539 \\
    & MLP & \textcolor{rhoNegRed}{\textbf{-0.0302}} & 0.8810 \\
    \bottomrule
    \end{tabular} & \begin{tabular}{l l c c}
    \toprule
    \textbf{Met.} & \textbf{Stg.} & \textbf{$\rho$} & \textbf{$p$} \\
    \midrule
    \rowcolor{sigLightGreen} Curv & Pre & \textcolor{rhoPosBlue}{\textbf{0.5378}} & 0.0038 \\
    & Att & \textcolor{rhoPosBlue}{\textbf{0.2458}} & 0.2164 \\
    \rowcolor{sigLightGreen}  & Res & \textcolor{rhoPosBlue}{\textbf{0.5195}} & 0.0055 \\
    & MLP & \textcolor{rhoPosBlue}{\textbf{0.3555}} & 0.0688 \\
    \bottomrule
    \end{tabular} \\[-0.3ex]
    \noalign{\vskip 1.5em}
    \begin{tabular}{l l c c}
    \toprule
    \textbf{Met.} & \textbf{Stg.} & \textbf{$\rho$} & \textbf{$p$} \\
    \midrule
    \rowcolor{sigLightGreen} ERank & Pre & \textcolor{rhoPosBlue}{\textbf{0.4205}} & 0.0290 \\
    & Att & \textcolor{rhoPosBlue}{\textbf{0.1154}} & 0.5664 \\
    \rowcolor{sigLightGreen}  & Res & \textcolor{rhoPosBlue}{\textbf{0.4550}} & 0.0171 \\
    & MLP & \textcolor{rhoPosBlue}{\textbf{0.3585}} & 0.0663 \\
    \bottomrule
    \end{tabular} & \begin{tabular}{l l c c}
    \toprule
    \textbf{Met.} & \textbf{Stg.} & \textbf{$\rho$} & \textbf{$p$} \\
    \midrule
    \rowcolor{sigLightGreen} LiDAR & Pre & \textcolor{rhoPosBlue}{\textbf{0.4294}} & 0.0254 \\
    & Att & \textcolor{rhoPosBlue}{\textbf{0.1594}} & 0.4271 \\
    \rowcolor{sigLightGreen}  & Res & \textcolor{rhoPosBlue}{\textbf{0.4120}} & 0.0327 \\
    & MLP & \textcolor{rhoPosBlue}{\textbf{0.2745}} & 0.1658 \\
    \bottomrule
    \end{tabular} & \begin{tabular}{l l c c}
    \toprule
    \textbf{Met.} & \textbf{Stg.} & \textbf{$\rho$} & \textbf{$p$} \\
    \midrule
    \rowcolor{sigLightGreen} Entr & Pre & \textcolor{rhoPosBlue}{\textbf{0.4184}} & 0.0299 \\
    & Att & \textcolor{rhoPosBlue}{\textbf{0.1191}} & 0.5541 \\
    \rowcolor{sigLightGreen}  & Res & \textcolor{rhoPosBlue}{\textbf{0.4483}} & 0.0190 \\
    & MLP & \textcolor{rhoPosBlue}{\textbf{0.3585}} & 0.0663 \\
    \bottomrule
    \end{tabular} \\
    \end{tabular}
    }

    \caption{\textbf{Gemma -- PubMed (metrics–performance drop correlations).}}
    \label{tab:layer_metric_ablation_correlations_gemma7b_pubmed}
    \end{table*}

\begin{table*}[t]
    \centering
    \small

    \resizebox{\textwidth}{!}{%
    \begin{tabular}{ccc}
    \begin{tabular}{l l c c}
    \toprule
    \textbf{Met.} & \textbf{Stg.} & \textbf{$\rho$} & \textbf{$p$} \\
    \midrule
    \rowcolor{sigLightGreen} DiME & Pre & \textcolor{rhoPosBlue}{\textbf{0.4869}} & 0.0100 \\
    \rowcolor{sigLightGreen}  & Att & \textcolor{rhoPosBlue}{\textbf{0.6295}} & 0.0004 \\
    \rowcolor{sigLightGreen}  & Res & \textcolor{rhoPosBlue}{\textbf{0.5382}} & 0.0038 \\
    \rowcolor{sigLightGreen}  & MLP & \textcolor{rhoPosBlue}{\textbf{0.5645}} & 0.0022 \\
    \bottomrule
    \end{tabular} & \begin{tabular}{l l c c}
    \toprule
    \textbf{Met.} & \textbf{Stg.} & \textbf{$\rho$} & \textbf{$p$} \\
    \midrule
    Info & Pre & \textcolor{rhoNegRed}{\textbf{-0.3447}} & 0.0783 \\
    & Att & \textcolor{rhoNegRed}{\textbf{-0.1084}} & 0.5905 \\
    & Res & \textcolor{rhoNegRed}{\textbf{-0.2742}} & 0.1664 \\
    & MLP & \textcolor{rhoNegRed}{\textbf{-0.2745}} & 0.1659 \\
    \bottomrule
    \end{tabular} & \begin{tabular}{l l c c}
    \toprule
    \textbf{Met.} & \textbf{Stg.} & \textbf{$\rho$} & \textbf{$p$} \\
    \midrule
    \rowcolor{sigLightGreen} Curv & Pre & \textcolor{rhoPosBlue}{\textbf{0.6252}} & 0.0005 \\
    & Att & \textcolor{rhoPosBlue}{\textbf{0.2436}} & 0.2207 \\
    \rowcolor{sigLightGreen}  & Res & \textcolor{rhoPosBlue}{\textbf{0.6088}} & 0.0008 \\
    \rowcolor{sigLightGreen}  & MLP & \textcolor{rhoPosBlue}{\textbf{0.5266}} & 0.0048 \\
    \bottomrule
    \end{tabular} \\[-0.3ex]
    \noalign{\vskip 1.5em}
    \begin{tabular}{l l c c}
    \toprule
    \textbf{Met.} & \textbf{Stg.} & \textbf{$\rho$} & \textbf{$p$} \\
    \midrule
    ERank & Pre & \textcolor{rhoPosBlue}{\textbf{0.2549}} & 0.1994 \\
    & Att & \textcolor{rhoPosBlue}{\textbf{0.2934}} & 0.1375 \\
    & Res & \textcolor{rhoPosBlue}{\textbf{0.2552}} & 0.1988 \\
    \rowcolor{sigLightGreen}  & MLP & \textcolor{rhoPosBlue}{\textbf{0.5187}} & 0.0056 \\
    \bottomrule
    \end{tabular} & \begin{tabular}{l l c c}
    \toprule
    \textbf{Met.} & \textbf{Stg.} & \textbf{$\rho$} & \textbf{$p$} \\
    \midrule
    LiDAR & Pre & \textcolor{rhoPosBlue}{\textbf{0.3792}} & 0.0511 \\
    & Att & \textcolor{rhoPosBlue}{\textbf{0.3566}} & 0.0679 \\
    \rowcolor{sigLightGreen}  & Res & \textcolor{rhoPosBlue}{\textbf{0.3831}} & 0.0485 \\
    \rowcolor{sigLightGreen}  & MLP & \textcolor{rhoPosBlue}{\textbf{0.4057}} & 0.0357 \\
    \bottomrule
    \end{tabular} & \begin{tabular}{l l c c}
    \toprule
    \textbf{Met.} & \textbf{Stg.} & \textbf{$\rho$} & \textbf{$p$} \\
    \midrule
    Entr & Pre & \textcolor{rhoPosBlue}{\textbf{0.2549}} & 0.1994 \\
    & Att & \textcolor{rhoPosBlue}{\textbf{0.2934}} & 0.1375 \\
    & Res & \textcolor{rhoPosBlue}{\textbf{0.2552}} & 0.1988 \\
    \rowcolor{sigLightGreen}  & MLP & \textcolor{rhoPosBlue}{\textbf{0.5187}} & 0.0056 \\
    \bottomrule
    \end{tabular} \\
    \end{tabular}
    }

    \caption{\textbf{Gemma -- GovReport (metrics–performance drop correlations).}}
    \label{tab:layer_metric_ablation_correlations_gemma7b_govreport}
\end{table*}

\begin{table*}[t]
    \centering
    \small
    \setlength{\tabcolsep}{4pt}

    \resizebox{\textwidth}{!}{%
    \begin{minipage}{\textwidth}
        \centering

        \begin{minipage}[t]{0.32\textwidth}
        \centering
        \begin{tabular}{l l c c}
        \toprule
        \textbf{Met.} & \textbf{Stg.} & \textbf{$\rho$} & \textbf{$p$} \\
        \midrule
        \rowcolor{sigLightGreen} DiME & Pre & \textcolor{rhoPosBlue}{\textbf{0.4331}} & 0.0240 \\
         & Att & \textcolor{rhoPosBlue}{\textbf{0.1814}} & 0.3651 \\
        \rowcolor{sigLightGreen} & Res & \textcolor{rhoPosBlue}{\textbf{0.4117}} & 0.0329 \\
        \rowcolor{sigLightGreen} & MLP & \textcolor{rhoPosBlue}{\textbf{0.3910}} & 0.0438 \\
        \bottomrule
        \end{tabular}
        \end{minipage}\hfill
        \begin{minipage}[t]{0.32\textwidth}
        \centering
        \begin{tabular}{l l c c}
        \toprule
        \textbf{Met.} & \textbf{Stg.} & \textbf{$\rho$} & \textbf{$p$} \\
        \midrule
        Info & Pre & \textcolor{rhoNegRed}{\textbf{-0.2572}} & 0.1953 \\
        \rowcolor{sigLightGreen} & Att & \textcolor{rhoNegRed}{\textbf{-0.4728}} & 0.0128 \\
        \rowcolor{sigLightGreen} & Res & \textcolor{rhoNegRed}{\textbf{-0.3824}} & 0.0490 \\
        \rowcolor{sigLightGreen} & MLP & \textcolor{rhoNegRed}{\textbf{-0.3934}} & 0.0423 \\
        \bottomrule
        \end{tabular}
        \end{minipage}\hfill
        \begin{minipage}[t]{0.32\textwidth}
        \centering
        \begin{tabular}{l l c c}
        \toprule
        \textbf{Met.} & \textbf{Stg.} & \textbf{$\rho$} & \textbf{$p$} \\
        \midrule
        \rowcolor{sigLightGreen} Curv & Pre & \textcolor{rhoPosBlue}{\textbf{0.4307}} & 0.0249 \\
         & Att & \textcolor{rhoPosBlue}{\textbf{0.2737}} & 0.1672 \\
        \rowcolor{sigLightGreen} & Res & \textcolor{rhoPosBlue}{\textbf{0.4484}} & 0.0190 \\
         & MLP & \textcolor{rhoPosBlue}{\textbf{0.2828}} & 0.1529 \\
        \bottomrule
        \end{tabular}
        \end{minipage}

        \vspace{1.2em}

        \begin{minipage}[t]{0.32\textwidth}
        \centering
        \begin{tabular}{l l c c}
        \toprule
        \textbf{Met.} & \textbf{Stg.} & \textbf{$\rho$} & \textbf{$p$} \\
        \midrule
        ERank & Pre & \textcolor{rhoPosBlue}{\textbf{0.3195}} & 0.1043 \\
         & Att & \textcolor{rhoPosBlue}{\textbf{0.2108}} & 0.2913 \\
        \rowcolor{sigLightGreen} & Res & \textcolor{rhoPosBlue}{\textbf{0.3910}} & 0.0438 \\
         & MLP & \textcolor{rhoPosBlue}{\textbf{0.2260}} & 0.2570 \\
        \bottomrule
        \end{tabular}
        \end{minipage}\hfill
        \begin{minipage}[t]{0.32\textwidth}
        \centering
        \begin{tabular}{l l c c}
        \toprule
        \textbf{Met.} & \textbf{Stg.} & \textbf{$\rho$} & \textbf{$p$} \\
        \midrule
        LiDAR & Pre & \textcolor{rhoPosBlue}{\textbf{0.2731}} & 0.1682 \\
         & Att & \textcolor{rhoNegRed}{\textbf{-0.0006}} & 0.9976 \\
         & Res & \textcolor{rhoPosBlue}{\textbf{0.3018}} & 0.1261 \\
         & MLP & \textcolor{rhoPosBlue}{\textbf{0.1546}} & 0.4415 \\
        \bottomrule
        \end{tabular}
        \end{minipage}\hfill
        \begin{minipage}[t]{0.32\textwidth}
        \centering
        \begin{tabular}{l l c c}
        \toprule
        \textbf{Met.} & \textbf{Stg.} & \textbf{$\rho$} & \textbf{$p$} \\
        \midrule
        Entr & Pre & \textcolor{rhoPosBlue}{\textbf{0.3195}} & 0.1043 \\
         & Att & \textcolor{rhoPosBlue}{\textbf{0.2108}} & 0.2913 \\
        \rowcolor{sigLightGreen} & Res & \textcolor{rhoPosBlue}{\textbf{0.3910}} & 0.0438 \\
         & MLP & \textcolor{rhoPosBlue}{\textbf{0.2260}} & 0.2570 \\
        \bottomrule
        \end{tabular}
        \end{minipage}
    \end{minipage}%
    }

    \caption{\textbf{Gemma -- arXiv (metrics–performance drop correlations).}}
    \label{tab:layer_metric_ablation_correlations_gemma7b_arxiv}
\end{table*}

\begin{table*}[t]
    \centering
    \small

    \resizebox{\textwidth}{!}{%
    \begin{tabular}{ccc}
    \begin{tabular}{l l c c}
    \toprule
    \textbf{Met.} & \textbf{Stg.} & \textbf{$\rho$} & \textbf{$p$} \\
    \midrule
    \rowcolor{sigLightGreen} DiME & Pre & \textcolor{rhoPosBlue}{\textbf{0.4198}} & 0.0293 \\
    \rowcolor{sigLightGreen}  & Att & \textcolor{rhoPosBlue}{\textbf{0.6042}} & 0.0008 \\
    \rowcolor{sigLightGreen}  & Res & \textcolor{rhoPosBlue}{\textbf{0.4009}} & 0.0383 \\
    \rowcolor{sigLightGreen}  & MLP & \textcolor{rhoPosBlue}{\textbf{0.5153}} & 0.0059 \\
    \bottomrule
    \end{tabular} & \begin{tabular}{l l c c}
    \toprule
    \textbf{Met.} & \textbf{Stg.} & \textbf{$\rho$} & \textbf{$p$} \\
    \midrule
    \rowcolor{sigLightGreen} Info & Pre & \textcolor{rhoNegRed}{\textbf{-0.4644}} & 0.0147 \\
    \rowcolor{sigLightGreen}  & Att & \textcolor{rhoNegRed}{\textbf{-0.5327}} & 0.0042 \\
    \rowcolor{sigLightGreen}  & Res & \textcolor{rhoNegRed}{\textbf{-0.4479}} & 0.0191 \\
    \rowcolor{sigLightGreen}  & MLP & \textcolor{rhoNegRed}{\textbf{-0.4677}} & 0.0139 \\
    \bottomrule
    \end{tabular} & \begin{tabular}{l l c c}
    \toprule
    \textbf{Met.} & \textbf{Stg.} & \textbf{$\rho$} & \textbf{$p$} \\
    \midrule
    \rowcolor{sigLightGreen} Curv & Pre & \textcolor{rhoPosBlue}{\textbf{0.6891}} & 0.0001 \\
    & Att & \textcolor{rhoPosBlue}{\textbf{0.3499}} & 0.0736 \\
    \rowcolor{sigLightGreen}  & Res & \textcolor{rhoPosBlue}{\textbf{0.7526}} & 0.0000 \\
    \rowcolor{sigLightGreen}  & MLP & \textcolor{rhoPosBlue}{\textbf{0.7690}} & 0.0000 \\
    \bottomrule
    \end{tabular} \\[-0.3ex]
    \noalign{\vskip 1.5em}
    \begin{tabular}{l l c c}
    \toprule
    \textbf{Met.} & \textbf{Stg.} & \textbf{$\rho$} & \textbf{$p$} \\
    \midrule
    ERank & Pre & \textcolor{rhoPosBlue}{\textbf{0.2568}} & 0.1961 \\
    & Att & \textcolor{rhoPosBlue}{\textbf{0.3374}} & 0.0853 \\
    & Res & \textcolor{rhoPosBlue}{\textbf{0.2781}} & 0.1601 \\
    \rowcolor{sigLightGreen}  & MLP & \textcolor{rhoPosBlue}{\textbf{0.7431}} & 0.0000 \\
    \bottomrule
    \end{tabular} & \begin{tabular}{l l c c}
    \toprule
    \textbf{Met.} & \textbf{Stg.} & \textbf{$\rho$} & \textbf{$p$} \\
    \midrule
    LiDAR & Pre & \textcolor{rhoPosBlue}{\textbf{0.2268}} & 0.2552 \\
    & Att & \textcolor{rhoNegRed}{\textbf{-0.2629}} & 0.1853 \\
    & Res & \textcolor{rhoPosBlue}{\textbf{0.2027}} & 0.3105 \\
    & MLP & \textcolor{rhoNegRed}{\textbf{-0.1072}} & 0.5947 \\
    \bottomrule
    \end{tabular} & \begin{tabular}{l l c c}
    \toprule
    \textbf{Met.} & \textbf{Stg.} & \textbf{$\rho$} & \textbf{$p$} \\
    \midrule
    Entr & Pre & \textcolor{rhoPosBlue}{\textbf{0.2568}} & 0.1961 \\
    & Att & \textcolor{rhoPosBlue}{\textbf{0.3404}} & 0.0823 \\
    & Res & \textcolor{rhoPosBlue}{\textbf{0.2781}} & 0.1601 \\
    \rowcolor{sigLightGreen}  & MLP & \textcolor{rhoPosBlue}{\textbf{0.7431}} & 0.0000 \\
    \bottomrule
    \end{tabular} \\
    \end{tabular}
    }

    \caption{\textbf{Gemma -- LegalCase (metrics–performance drop correlations).}}
    \label{tab:layer_metric_ablation_correlations_gemma7b_legalsumm}
\end{table*}

\begin{table*}[t]
    \centering
    \small

    \resizebox{\textwidth}{!}{%
    \begin{tabular}{ccc}
    \begin{tabular}{l l c c}
   \toprule
   \textbf{Met.} & \textbf{Stg.} & \textbf{$\rho$} & \textbf{$p$} \\
   \midrule
   DiME & Pre & \textcolor{rhoPosBlue}{\textbf{0.2545}} & 0.1671 \\
   & Att & \textcolor{rhoNegRed}{\textbf{-0.1105}} & 0.5540 \\
   & Res & \textcolor{rhoPosBlue}{\textbf{0.1053}} & 0.5731 \\
   & MLP & \textcolor{rhoPosBlue}{\textbf{0.1345}} & 0.4707 \\
   \bottomrule
   \end{tabular} & \begin{tabular}{l l c c}
    \toprule
    \textbf{Met.} & \textbf{Stg.} & \textbf{$\rho$} & \textbf{$p$} \\
    \midrule
    Info & Pre & \textcolor{rhoNegRed}{\textbf{-0.0550}} & 0.7687 \\
    \rowcolor{sigLightGreen}  & Att & \textcolor{rhoNegRed}{\textbf{-0.5053}} & 0.0037 \\
    \rowcolor{sigLightGreen}  & Res & \textcolor{rhoNegRed}{\textbf{-0.3658}} & 0.0430 \\
    \rowcolor{sigLightGreen}  & MLP & \textcolor{rhoNegRed}{\textbf{-0.4051}} & 0.0238 \\
    \bottomrule
    \end{tabular} & \begin{tabular}{l l c c}
   \toprule
   \textbf{Met.} & \textbf{Stg.} & \textbf{$\rho$} & \textbf{$p$} \\
   \midrule
   Curv & Pre & \textcolor{rhoNegRed}{\textbf{-0.0331}} & 0.8598 \\
   & Att & \textcolor{rhoPosBlue}{\textbf{0.2061}} & 0.2661 \\
   & Res & \textcolor{rhoPosBlue}{\textbf{0.0228}} & 0.9032 \\
   & MLP & \textcolor{rhoPosBlue}{\textbf{0.0754}} & 0.6868 \\
   \bottomrule
   \end{tabular} \\[-0.3ex]
   \noalign{\vskip 1.5em}
    \begin{tabular}{l l c c}
   \toprule
   \textbf{Met.} & \textbf{Stg.} & \textbf{$\rho$} & \textbf{$p$} \\
   \midrule
   ERank & Pre & \textcolor{rhoNegRed}{\textbf{-0.0357}} & 0.8488 \\
   & Att & \textcolor{rhoNegRed}{\textbf{-0.0919}} & 0.6228 \\
   & Res & \textcolor{rhoNegRed}{\textbf{-0.0391}} & 0.8345 \\
   & MLP & \textcolor{rhoPosBlue}{\textbf{0.0288}} & 0.8776 \\
   \bottomrule
   \end{tabular} & \begin{tabular}{l l c c}
   \toprule
   \textbf{Met.} & \textbf{Stg.} & \textbf{$\rho$} & \textbf{$p$} \\
   \midrule
   LiDAR & Pre & \textcolor{rhoPosBlue}{\textbf{0.1970}} & 0.2881 \\
   & Att & \textcolor{rhoNegRed}{\textbf{-0.1214}} & 0.5154 \\
   & Res & \textcolor{rhoPosBlue}{\textbf{0.1784}} & 0.3368 \\
   & MLP & \textcolor{rhoPosBlue}{\textbf{0.1121}} & 0.5482 \\
   \bottomrule
   \end{tabular} & \begin{tabular}{l l c c}
   \toprule
   \textbf{Met.} & \textbf{Stg.} & \textbf{$\rho$} & \textbf{$p$} \\
   \midrule
   Entr & Pre & \textcolor{rhoNegRed}{\textbf{-0.0357}} & 0.8488 \\
   & Att & \textcolor{rhoNegRed}{\textbf{-0.0919}} & 0.6228 \\
   & Res & \textcolor{rhoNegRed}{\textbf{-0.0383}} & 0.8379 \\
   & MLP & \textcolor{rhoPosBlue}{\textbf{0.0373}} & 0.8421 \\
   \bottomrule
   \end{tabular} \\
    \end{tabular}
    }

    \caption{\textbf{LLaMA -- PubMed (metrics–performance drop correlations).}}
    \label{tab:layer_metric_ablation_correlations_llama31_pubmed}
    \end{table*}

\begin{table*}[t]
    \centering
    \small

    \resizebox{\textwidth}{!}{%
    \begin{tabular}{ccc}
    \begin{tabular}{l l c c}
   \toprule
   \textbf{Met.} & \textbf{Stg.} & \textbf{$\rho$} & \textbf{$p$} \\
   \midrule
   DiME & Pre & \textcolor{rhoPosBlue}{\textbf{0.0012}} & 0.9948 \\
   & Att & \textcolor{rhoNegRed}{\textbf{-0.2282}} & 0.2169 \\
   & Res & \textcolor{rhoNegRed}{\textbf{-0.0169}} & 0.9280 \\
   & MLP & \textcolor{rhoNegRed}{\textbf{-0.1294}} & 0.4877 \\
   \bottomrule
   \end{tabular} & \begin{tabular}{l l c c}
   \toprule
   \textbf{Met.} & \textbf{Stg.} & \textbf{$\rho$} & \textbf{$p$} \\
   \midrule
   Info & Pre & \textcolor{rhoPosBlue}{\textbf{0.3190}} & 0.0803 \\
   & Att & \textcolor{rhoNegRed}{\textbf{-0.0903}} & 0.6289 \\
   & Res & \textcolor{rhoPosBlue}{\textbf{0.1444}} & 0.4385 \\
   & MLP & \textcolor{rhoNegRed}{\textbf{-0.0089}} & 0.9622 \\
   \bottomrule
   \end{tabular} & \begin{tabular}{l l c c}
   \toprule
   \textbf{Met.} & \textbf{Stg.} & \textbf{$\rho$} & \textbf{$p$} \\
   \midrule
   Curv & Pre & \textcolor{rhoNegRed}{\textbf{-0.2185}} & 0.2375 \\
   & Att & \textcolor{rhoNegRed}{\textbf{-0.0085}} & 0.9639 \\
   & Res & \textcolor{rhoNegRed}{\textbf{-0.1940}} & 0.2958 \\
   & MLP & \textcolor{rhoNegRed}{\textbf{-0.1629}} & 0.3812 \\
   \bottomrule
   \end{tabular} \\[-0.3ex]
   \noalign{\vskip 1.5em}
    \begin{tabular}{l l c c}
    \toprule
    \textbf{Met.} & \textbf{Stg.} & \textbf{$\rho$} & \textbf{$p$} \\
    \midrule
    \rowcolor{sigLightGreen} ERank & Pre & \textcolor{rhoNegRed}{\textbf{-0.3565}} & 0.0490 \\
    & Att & \textcolor{rhoNegRed}{\textbf{-0.1746}} & 0.3475 \\
    \rowcolor{sigLightGreen}  & Res & \textcolor{rhoNegRed}{\textbf{-0.3601}} & 0.0466 \\
    & MLP & \textcolor{rhoNegRed}{\textbf{-0.2226}} & 0.2288 \\
    \bottomrule
    \end{tabular} & \begin{tabular}{l l c c}
   \toprule
   \textbf{Met.} & \textbf{Stg.} & \textbf{$\rho$} & \textbf{$p$} \\
   \midrule
   LiDAR & Pre & \textcolor{rhoNegRed}{\textbf{-0.0742}} & 0.6916 \\
   & Att & \textcolor{rhoNegRed}{\textbf{-0.2621}} & 0.1543 \\
   & Res & \textcolor{rhoNegRed}{\textbf{-0.0984}} & 0.5985 \\
   & MLP & \textcolor{rhoNegRed}{\textbf{-0.1137}} & 0.5425 \\
   \bottomrule
   \end{tabular} & \begin{tabular}{l l c c}
    \toprule
    \textbf{Met.} & \textbf{Stg.} & \textbf{$\rho$} & \textbf{$p$} \\
    \midrule
    \rowcolor{sigLightGreen} Entr & Pre & \textcolor{rhoNegRed}{\textbf{-0.3565}} & 0.0490 \\
    & Att & \textcolor{rhoNegRed}{\textbf{-0.1746}} & 0.3475 \\
    \rowcolor{sigLightGreen}  & Res & \textcolor{rhoNegRed}{\textbf{-0.3601}} & 0.0466 \\
    & MLP & \textcolor{rhoNegRed}{\textbf{-0.2206}} & 0.2331 \\
    \bottomrule
    \end{tabular} \\
    \end{tabular}
    }

    \caption{\textbf{LLaMA -- GovReport (metrics–performance drop correlations).}}
    \label{tab:layer_metric_ablation_correlations_llama31_govreport}
    \end{table*}

\begin{table*}[t]
    \centering
    \small

    \resizebox{\textwidth}{!}{%
    \begin{tabular}{ccc}
    \begin{tabular}{l l c c}
   \toprule
   \textbf{Met.} & \textbf{Stg.} & \textbf{$\rho$} & \textbf{$p$} \\
   \midrule
   DiME & Pre & \textcolor{rhoPosBlue}{\textbf{0.2599}} & 0.1579 \\
   & Att & \textcolor{rhoNegRed}{\textbf{-0.0661}} & 0.7237 \\
   & Res & \textcolor{rhoPosBlue}{\textbf{0.1218}} & 0.5140 \\
   & MLP & \textcolor{rhoPosBlue}{\textbf{0.1422}} & 0.4456 \\
   \bottomrule
   \end{tabular} & \begin{tabular}{l l c c}
    \toprule
    \textbf{Met.} & \textbf{Stg.} & \textbf{$\rho$} & \textbf{$p$} \\
    \midrule
    Info & Pre & \textcolor{rhoNegRed}{\textbf{-0.2168}} & 0.2415 \\
    \rowcolor{sigLightGreen}  & Att & \textcolor{rhoNegRed}{\textbf{-0.5053}} & 0.0037 \\
    \rowcolor{sigLightGreen}  & Res & \textcolor{rhoNegRed}{\textbf{-0.4029}} & 0.0246 \\
    \rowcolor{sigLightGreen}  & MLP & \textcolor{rhoNegRed}{\textbf{-0.4180}} & 0.0193 \\
    \bottomrule
    \end{tabular} & \begin{tabular}{l l c c}
   \toprule
   \textbf{Met.} & \textbf{Stg.} & \textbf{$\rho$} & \textbf{$p$} \\
   \midrule
   Curv & Pre & \textcolor{rhoNegRed}{\textbf{-0.0667}} & 0.7213 \\
   & Att & \textcolor{rhoNegRed}{\textbf{-0.1059}} & 0.5709 \\
   & Res & \textcolor{rhoNegRed}{\textbf{-0.0645}} & 0.7302 \\
   & MLP & \textcolor{rhoNegRed}{\textbf{-0.0133}} & 0.9434 \\
   \bottomrule
   \end{tabular} \\[-0.3ex]
   \noalign{\vskip 1.5em}
    \begin{tabular}{l l c c}
   \toprule
   \textbf{Met.} & \textbf{Stg.} & \textbf{$\rho$} & \textbf{$p$} \\
   \midrule
   ERank & Pre & \textcolor{rhoNegRed}{\textbf{-0.0718}} & 0.7012 \\
   & Att & \textcolor{rhoNegRed}{\textbf{-0.2299}} & 0.2135 \\
   & Res & \textcolor{rhoNegRed}{\textbf{-0.0829}} & 0.6576 \\
   & MLP & \textcolor{rhoNegRed}{\textbf{-0.1034}} & 0.5798 \\
   \bottomrule
   \end{tabular} & \begin{tabular}{l l c c}
   \toprule
   \textbf{Met.} & \textbf{Stg.} & \textbf{$\rho$} & \textbf{$p$} \\
   \midrule
   LiDAR & Pre & \textcolor{rhoPosBlue}{\textbf{0.1974}} & 0.2871 \\
   & Att & \textcolor{rhoNegRed}{\textbf{-0.0177}} & 0.9245 \\
   & Res & \textcolor{rhoPosBlue}{\textbf{0.1861}} & 0.3161 \\
   & MLP & \textcolor{rhoPosBlue}{\textbf{0.1317}} & 0.4802 \\
   \bottomrule
   \end{tabular} & \begin{tabular}{l l c c}
   \toprule
   \textbf{Met.} & \textbf{Stg.} & \textbf{$\rho$} & \textbf{$p$} \\
   \midrule
   Entr & Pre & \textcolor{rhoNegRed}{\textbf{-0.0718}} & 0.7012 \\
   & Att & \textcolor{rhoNegRed}{\textbf{-0.2299}} & 0.2135 \\
   & Res & \textcolor{rhoNegRed}{\textbf{-0.0829}} & 0.6576 \\
   & MLP & \textcolor{rhoNegRed}{\textbf{-0.1006}} & 0.5902 \\
   \bottomrule
   \end{tabular} \\
    \end{tabular}
    }

    \caption{\textbf{LLaMA -- arXiv (metrics–performance drop correlations).}}
    \label{tab:layer_metric_ablation_correlations_llama31_arxiv}
    \end{table*}

\begin{table*}[t]
    \centering
    \small

    \resizebox{\textwidth}{!}{%
    \begin{tabular}{ccc}
    \begin{tabular}{l l c c}
   \toprule
   \textbf{Met.} & \textbf{Stg.} & \textbf{$\rho$} & \textbf{$p$} \\
   \midrule
   DiME & Pre & \textcolor{rhoNegRed}{\textbf{-0.2371}} & 0.1990 \\
   & Att & \textcolor{rhoPosBlue}{\textbf{0.0772}} & 0.6796 \\
   & Res & \textcolor{rhoNegRed}{\textbf{-0.0827}} & 0.6584 \\
   & MLP & \textcolor{rhoNegRed}{\textbf{-0.0446}} & 0.8118 \\
   \bottomrule
   \end{tabular} & \begin{tabular}{l l c c}
    \toprule
    \textbf{Met.} & \textbf{Stg.} & \textbf{$\rho$} & \textbf{$p$} \\
    \midrule
    \rowcolor{sigLightGreen} Info & Pre & \textcolor{rhoNegRed}{\textbf{-0.4011}} & 0.0253 \\
    & Att & \textcolor{rhoNegRed}{\textbf{-0.2634}} & 0.1523 \\
    \rowcolor{sigLightGreen}  & Res & \textcolor{rhoNegRed}{\textbf{-0.4047}} & 0.0239 \\
    & MLP & \textcolor{rhoNegRed}{\textbf{-0.3186}} & 0.0807 \\
    \bottomrule
    \end{tabular} & \begin{tabular}{l l c c}
    \toprule
    \textbf{Met.} & \textbf{Stg.} & \textbf{$\rho$} & \textbf{$p$} \\
    \midrule
    \rowcolor{sigLightGreen} Curv & Pre & \textcolor{rhoPosBlue}{\textbf{0.5580}} & 0.0011 \\
    \rowcolor{sigLightGreen}  & Att & \textcolor{rhoPosBlue}{\textbf{0.4019}} & 0.0250 \\
    \rowcolor{sigLightGreen}  & Res & \textcolor{rhoPosBlue}{\textbf{0.5604}} & 0.0010 \\
    \rowcolor{sigLightGreen}  & MLP & \textcolor{rhoPosBlue}{\textbf{0.5348}} & 0.0019 \\
    \bottomrule
    \end{tabular} \\[-0.3ex]
    \noalign{\vskip 1.5em}
    \begin{tabular}{l l c c}
    \toprule
    \textbf{Met.} & \textbf{Stg.} & \textbf{$\rho$} & \textbf{$p$} \\
    \midrule
    \rowcolor{sigLightGreen} ERank & Pre & \textcolor{rhoPosBlue}{\textbf{0.3636}} & 0.0444 \\
    \rowcolor{sigLightGreen}  & Att & \textcolor{rhoPosBlue}{\textbf{0.4783}} & 0.0065 \\
    \rowcolor{sigLightGreen}  & Res & \textcolor{rhoPosBlue}{\textbf{0.4098}} & 0.0221 \\
    \rowcolor{sigLightGreen}  & MLP & \textcolor{rhoPosBlue}{\textbf{0.5059}} & 0.0037 \\
    \bottomrule
    \end{tabular} & \begin{tabular}{l l c c}
   \toprule
   \textbf{Met.} & \textbf{Stg.} & \textbf{$\rho$} & \textbf{$p$} \\
   \midrule
   LiDAR & Pre & \textcolor{rhoNegRed}{\textbf{-0.1654}} & 0.3740 \\
   & Att & \textcolor{rhoNegRed}{\textbf{-0.1341}} & 0.4720 \\
   & Res & \textcolor{rhoNegRed}{\textbf{-0.1605}} & 0.3884 \\
   & MLP & \textcolor{rhoNegRed}{\textbf{-0.0970}} & 0.6037 \\
   \bottomrule
   \end{tabular} & \begin{tabular}{l l c c}
    \toprule
    \textbf{Met.} & \textbf{Stg.} & \textbf{$\rho$} & \textbf{$p$} \\
    \midrule
    \rowcolor{sigLightGreen} Entr & Pre & \textcolor{rhoPosBlue}{\textbf{0.3636}} & 0.0444 \\
    \rowcolor{sigLightGreen}  & Att & \textcolor{rhoPosBlue}{\textbf{0.4783}} & 0.0065 \\
    \rowcolor{sigLightGreen}  & Res & \textcolor{rhoPosBlue}{\textbf{0.4098}} & 0.0221 \\
    \rowcolor{sigLightGreen}  & MLP & \textcolor{rhoPosBlue}{\textbf{0.5102}} & 0.0034 \\
    \bottomrule
    \end{tabular} \\
    \end{tabular}
    }

    \caption{\textbf{LLaMA -- LegalCase (metrics–performance drop correlations).}}
    \label{tab:layer_metric_ablation_correlations_llama31_legalsumm}
    \end{table*}

\begin{table*}[t]
    \centering
    \small

    \resizebox{\textwidth}{!}{%
    \begin{tabular}{ccc}
    \begin{tabular}{l l c c}
   \toprule
   \textbf{Met.} & \textbf{Stg.} & \textbf{$\rho$} & \textbf{$p$} \\
   \midrule
   DiME & Pre & \textcolor{rhoPosBlue}{\textbf{0.2653}} & 0.1810 \\
   & Att & \textcolor{rhoNegRed}{\textbf{-0.0577}} & 0.7749 \\
   & Res & \textcolor{rhoPosBlue}{\textbf{0.1420}} & 0.4799 \\
   & MLP & \textcolor{rhoPosBlue}{\textbf{0.1612}} & 0.4218 \\
   \bottomrule
   \end{tabular} & \begin{tabular}{l l c c}
   \toprule
   \textbf{Met.} & \textbf{Stg.} & \textbf{$\rho$} & \textbf{$p$} \\
   \midrule
   \rowcolor{sigLightGreen} Info & Pre & \textcolor{rhoNegRed}{\textbf{-0.3935}} & 0.0423 \\
   & Att & \textcolor{rhoNegRed}{\textbf{-0.1924}} & 0.3364 \\
   & Res & \textcolor{rhoNegRed}{\textbf{-0.3524}} & 0.0714 \\
   & MLP & \textcolor{rhoNegRed}{\textbf{-0.3582}} & 0.0666 \\
   \bottomrule
   \end{tabular} & \begin{tabular}{l l c c}
   \toprule
   \textbf{Met.} & \textbf{Stg.} & \textbf{$\rho$} & \textbf{$p$} \\
   \midrule
   Curv & Pre & \textcolor{rhoNegRed}{\textbf{-0.1524}} & 0.4480 \\
   & Att & \textcolor{rhoNegRed}{\textbf{-0.2495}} & 0.2095 \\
   & Res & \textcolor{rhoNegRed}{\textbf{-0.0595}} & 0.7680 \\
   & MLP & \textcolor{rhoNegRed}{\textbf{-0.2125}} & 0.2872 \\
   \bottomrule
   \end{tabular} \\[-0.3ex]
   \noalign{\vskip 1.5em}
    \begin{tabular}{l l c c}
   \toprule
   \textbf{Met.} & \textbf{Stg.} & \textbf{$\rho$} & \textbf{$p$} \\
   \midrule
   ERank & Pre & \textcolor{rhoPosBlue}{\textbf{0.3634}} & 0.0625 \\
   & Att & \textcolor{rhoPosBlue}{\textbf{0.0168}} & 0.9337 \\
   & Res & \textcolor{rhoPosBlue}{\textbf{0.3487}} & 0.0747 \\
   & MLP & \textcolor{rhoNegRed}{\textbf{-0.2128}} & 0.2865 \\
   \bottomrule
   \end{tabular} & \begin{tabular}{l l c c}
   \toprule
   \textbf{Met.} & \textbf{Stg.} & \textbf{$\rho$} & \textbf{$p$} \\
   \midrule
   LiDAR & Pre & \textcolor{rhoPosBlue}{\textbf{0.3505}} & 0.0730 \\
   & Att & \textcolor{rhoPosBlue}{\textbf{0.3362}} & 0.0864 \\
   & Res & \textcolor{rhoPosBlue}{\textbf{0.3744}} & 0.0544 \\
   & MLP & \textcolor{rhoPosBlue}{\textbf{0.3646}} & 0.0615 \\
   \bottomrule
   \end{tabular} & \begin{tabular}{l l c c}
   \toprule
   \textbf{Met.} & \textbf{Stg.} & \textbf{$\rho$} & \textbf{$p$} \\
   \midrule
   Entr & Pre & \textcolor{rhoPosBlue}{\textbf{0.3634}} & 0.0625 \\
   & Att & \textcolor{rhoPosBlue}{\textbf{0.0150}} & 0.9410 \\
   & Res & \textcolor{rhoPosBlue}{\textbf{0.3487}} & 0.0747 \\
   & MLP & \textcolor{rhoNegRed}{\textbf{-0.2128}} & 0.2865 \\
   \bottomrule
   \end{tabular} \\
    \end{tabular}
    }

    \caption{\textbf{Qwen -- PubMed (metrics–performance drop correlations).}}
    \label{tab:layer_metric_ablation_correlations_qwen25_pubmed}
    \end{table*}

\begin{table*}[t]
    \centering
    \small

    \resizebox{\textwidth}{!}{%
    \begin{tabular}{ccc}
    \begin{tabular}{l l c c}
   \toprule
   \textbf{Met.} & \textbf{Stg.} & \textbf{$\rho$} & \textbf{$p$} \\
   \midrule
   DiME & Pre & \textcolor{rhoNegRed}{\textbf{-0.0366}} & 0.8560 \\
   & Att & \textcolor{rhoNegRed}{\textbf{-0.0901}} & 0.6551 \\
   & Res & \textcolor{rhoNegRed}{\textbf{-0.2482}} & 0.2119 \\
   & MLP & \textcolor{rhoNegRed}{\textbf{-0.0464}} & 0.8182 \\
   \bottomrule
   \end{tabular} & \begin{tabular}{l l c c}
    \toprule
    \textbf{Met.} & \textbf{Stg.} & \textbf{$\rho$} & \textbf{$p$} \\
    \midrule
    \rowcolor{sigLightGreen} Info & Pre & \textcolor{rhoNegRed}{\textbf{-0.3859}} & 0.0468 \\
    & Att & \textcolor{rhoNegRed}{\textbf{-0.2513}} & 0.2062 \\
    \rowcolor{sigLightGreen}  & Res & \textcolor{rhoNegRed}{\textbf{-0.4463}} & 0.0196 \\
    & MLP & \textcolor{rhoNegRed}{\textbf{-0.1307}} & 0.5159 \\
    \bottomrule
    \end{tabular} & \begin{tabular}{l l c c}
    \toprule
    \textbf{Met.} & \textbf{Stg.} & \textbf{$\rho$} & \textbf{$p$} \\
    \midrule
    Curv & Pre & \textcolor{rhoNegRed}{\textbf{-0.0885}} & 0.6606 \\
    & Att & \textcolor{rhoNegRed}{\textbf{-0.3044}} & 0.1227 \\
    & Res & \textcolor{rhoNegRed}{\textbf{-0.3212}} & 0.1024 \\
    \rowcolor{sigLightGreen}  & MLP & \textcolor{rhoNegRed}{\textbf{-0.4045}} & 0.0364 \\
    \bottomrule
    \end{tabular} \\[-0.3ex]
    \noalign{\vskip 1.5em}
    \begin{tabular}{l l c c}
    \toprule
    \textbf{Met.} & \textbf{Stg.} & \textbf{$\rho$} & \textbf{$p$} \\
    \midrule
    \rowcolor{sigLightGreen} ERank & Pre & \textcolor{rhoPosBlue}{\textbf{0.4937}} & 0.0089 \\
    & Att & \textcolor{rhoNegRed}{\textbf{-0.0119}} & 0.9530 \\
    \rowcolor{sigLightGreen}  & Res & \textcolor{rhoPosBlue}{\textbf{0.4854}} & 0.0103 \\
    & MLP & \textcolor{rhoNegRed}{\textbf{-0.1679}} & 0.4025 \\
    \bottomrule
    \end{tabular} & \begin{tabular}{l l c c}
    \toprule
    \textbf{Met.} & \textbf{Stg.} & \textbf{$\rho$} & \textbf{$p$} \\
    \midrule
    LiDAR & Pre & \textcolor{rhoPosBlue}{\textbf{0.3389}} & 0.0838 \\
    & Att & \textcolor{rhoPosBlue}{\textbf{0.3697}} & 0.0577 \\
    \rowcolor{sigLightGreen}  & Res & \textcolor{rhoPosBlue}{\textbf{0.3911}} & 0.0437 \\
    & MLP & \textcolor{rhoPosBlue}{\textbf{0.2097}} & 0.2937 \\
    \bottomrule
    \end{tabular} & \begin{tabular}{l l c c}
    \toprule
    \textbf{Met.} & \textbf{Stg.} & \textbf{$\rho$} & \textbf{$p$} \\
    \midrule
    \rowcolor{sigLightGreen} Entr & Pre & \textcolor{rhoPosBlue}{\textbf{0.5028}} & 0.0075 \\
    & Att & \textcolor{rhoNegRed}{\textbf{-0.0119}} & 0.9530 \\
    \rowcolor{sigLightGreen}  & Res & \textcolor{rhoPosBlue}{\textbf{0.4854}} & 0.0103 \\
    & MLP & \textcolor{rhoNegRed}{\textbf{-0.1557}} & 0.4380 \\
    \bottomrule
    \end{tabular} \\
    \end{tabular}
    }

    \caption{\textbf{Qwen -- GovReport (metrics–performance drop correlations).}}
    \label{tab:layer_metric_ablation_correlations_qwen25_govreport}
    \end{table*}

\begin{table*}[t]
    \centering
    \small

    \resizebox{\textwidth}{!}{%
    \begin{tabular}{ccc}
    \begin{tabular}{l l c c}
    \toprule
    \textbf{Met.} & \textbf{Stg.} & \textbf{$\rho$} & \textbf{$p$} \\
    \midrule
    DiME & Pre & \textcolor{rhoPosBlue}{\textbf{0.2144}} & 0.2829 \\
    & Att & \textcolor{rhoNegRed}{\textbf{-0.0195}} & 0.9229 \\
    & Res & \textcolor{rhoPosBlue}{\textbf{0.2024}} & 0.3114 \\
    \rowcolor{sigLightGreen}  & MLP & \textcolor{rhoPosBlue}{\textbf{0.4932}} & 0.0089 \\
    \bottomrule
    \end{tabular} & \begin{tabular}{l l c c}
    \toprule
    \textbf{Met.} & \textbf{Stg.} & \textbf{$\rho$} & \textbf{$p$} \\
    \midrule
    Info & Pre & \textcolor{rhoNegRed}{\textbf{-0.3393}} & 0.0834 \\
    & Att & \textcolor{rhoNegRed}{\textbf{-0.1628}} & 0.4172 \\
    & Res & \textcolor{rhoNegRed}{\textbf{-0.3735}} & 0.0550 \\
    \rowcolor{sigLightGreen}  & MLP & \textcolor{rhoNegRed}{\textbf{-0.4584}} & 0.0162 \\
    \bottomrule
    \end{tabular} & \begin{tabular}{l l c c}
    \toprule
    \textbf{Met.} & \textbf{Stg.} & \textbf{$\rho$} & \textbf{$p$} \\
    \midrule
    Curv & Pre & \textcolor{rhoNegRed}{\textbf{-0.2724}} & 0.1692 \\
    \rowcolor{sigLightGreen}  & Att & \textcolor{rhoNegRed}{\textbf{-0.5115}} & 0.0064 \\
    & Res & \textcolor{rhoNegRed}{\textbf{-0.3320}} & 0.0907 \\
    & MLP & \textcolor{rhoNegRed}{\textbf{-0.2489}} & 0.2106 \\
    \bottomrule
    \end{tabular} \\[-0.3ex]
    \noalign{\vskip 1.5em}
    \begin{tabular}{l l c c}
    \toprule
    \textbf{Met.} & \textbf{Stg.} & \textbf{$\rho$} & \textbf{$p$} \\
    \midrule
    \rowcolor{sigLightGreen} ERank & Pre & \textcolor{rhoPosBlue}{\textbf{0.4624}} & 0.0152 \\
    & Att & \textcolor{rhoNegRed}{\textbf{-0.1948}} & 0.3301 \\
    \rowcolor{sigLightGreen}  & Res & \textcolor{rhoPosBlue}{\textbf{0.4474}} & 0.0193 \\
    & MLP & \textcolor{rhoNegRed}{\textbf{-0.3280}} & 0.0949 \\
    \bottomrule
    \end{tabular} & \begin{tabular}{l l c c}
    \toprule
    \textbf{Met.} & \textbf{Stg.} & \textbf{$\rho$} & \textbf{$p$} \\
    \midrule
    \rowcolor{sigLightGreen} LiDAR & Pre & \textcolor{rhoPosBlue}{\textbf{0.4498}} & 0.0186 \\
    \rowcolor{sigLightGreen}  & Att & \textcolor{rhoPosBlue}{\textbf{0.4743}} & 0.0124 \\
    \rowcolor{sigLightGreen}  & Res & \textcolor{rhoPosBlue}{\textbf{0.4740}} & 0.0125 \\
    \rowcolor{sigLightGreen}  & MLP & \textcolor{rhoPosBlue}{\textbf{0.5256}} & 0.0049 \\
    \bottomrule
    \end{tabular} & \begin{tabular}{l l c c}
    \toprule
    \textbf{Met.} & \textbf{Stg.} & \textbf{$\rho$} & \textbf{$p$} \\
    \midrule
    \rowcolor{sigLightGreen} Entr & Pre & \textcolor{rhoPosBlue}{\textbf{0.4624}} & 0.0152 \\
    & Att & \textcolor{rhoNegRed}{\textbf{-0.2034}} & 0.3089 \\
    \rowcolor{sigLightGreen}  & Res & \textcolor{rhoPosBlue}{\textbf{0.4474}} & 0.0193 \\
    & MLP & \textcolor{rhoNegRed}{\textbf{-0.3280}} & 0.0949 \\
    \bottomrule
    \end{tabular} \\
    \end{tabular}
    }

    \caption{\textbf{Qwen -- arXiv (metrics–performance drop correlations).}}
    \label{tab:layer_metric_ablation_correlations_qwen25_arxiv}
    \end{table*}

\begin{table*}[t]
    \centering
    \small

    \resizebox{\textwidth}{!}{%
    \begin{tabular}{ccc}
    \begin{tabular}{l l c c}
    \toprule
    \textbf{Met.} & \textbf{Stg.} & \textbf{$\rho$} & \textbf{$p$} \\
    \midrule
    DiME & Pre & \textcolor{rhoNegRed}{\textbf{-0.0919}} & 0.6484 \\
    & Att & \textcolor{rhoNegRed}{\textbf{-0.1597}} & 0.4262 \\
    & Res & \textcolor{rhoNegRed}{\textbf{-0.2107}} & 0.2915 \\
    \rowcolor{sigLightGreen}  & MLP & \textcolor{rhoNegRed}{\textbf{-0.3853}} & 0.0471 \\
    \bottomrule
    \end{tabular} & \begin{tabular}{l l c c}
   \toprule
   \textbf{Met.} & \textbf{Stg.} & \textbf{$\rho$} & \textbf{$p$} \\
   \midrule
   Info & Pre & \textcolor{rhoNegRed}{\textbf{-0.2888}} & 0.1440 \\
   & Att & \textcolor{rhoPosBlue}{\textbf{0.1591}} & 0.4280 \\
   & Res & \textcolor{rhoPosBlue}{\textbf{0.0281}} & 0.8894 \\
   & MLP & \textcolor{rhoPosBlue}{\textbf{0.2247}} & 0.2598 \\
   \bottomrule
   \end{tabular} & \begin{tabular}{l l c c}
   \toprule
   \textbf{Met.} & \textbf{Stg.} & \textbf{$\rho$} & \textbf{$p$} \\
   \midrule
   Curv & Pre & \textcolor{rhoPosBlue}{\textbf{0.3142}} & 0.1105 \\
   & Att & \textcolor{rhoPosBlue}{\textbf{0.0244}} & 0.9037 \\
   & Res & \textcolor{rhoPosBlue}{\textbf{0.2959}} & 0.1340 \\
   & MLP & \textcolor{rhoPosBlue}{\textbf{0.1860}} & 0.3531 \\
   \bottomrule
   \end{tabular} \\[-0.3ex]
   \noalign{\vskip 1.5em}
    \begin{tabular}{l l c c}
    \toprule
    \textbf{Met.} & \textbf{Stg.} & \textbf{$\rho$} & \textbf{$p$} \\
    \midrule
    ERank & Pre & \textcolor{rhoNegRed}{\textbf{-0.3105}} & 0.1149 \\
    & Att & \textcolor{rhoPosBlue}{\textbf{0.2944}} & 0.1361 \\
    & Res & \textcolor{rhoNegRed}{\textbf{-0.2809}} & 0.1558 \\
    \rowcolor{sigLightGreen}  & MLP & \textcolor{rhoPosBlue}{\textbf{0.4046}} & 0.0363 \\
    \bottomrule
    \end{tabular} & \begin{tabular}{l l c c}
   \toprule
   \textbf{Met.} & \textbf{Stg.} & \textbf{$\rho$} & \textbf{$p$} \\
   \midrule
   LiDAR & Pre & \textcolor{rhoNegRed}{\textbf{-0.0177}} & 0.9301 \\
   & Att & \textcolor{rhoNegRed}{\textbf{-0.1545}} & 0.4416 \\
   & Res & \textcolor{rhoNegRed}{\textbf{-0.0693}} & 0.7312 \\
   & MLP & \textcolor{rhoNegRed}{\textbf{-0.2000}} & 0.3172 \\
   \bottomrule
   \end{tabular} & \begin{tabular}{l l c c}
    \toprule
    \textbf{Met.} & \textbf{Stg.} & \textbf{$\rho$} & \textbf{$p$} \\
    \midrule
    Entr & Pre & \textcolor{rhoNegRed}{\textbf{-0.3105}} & 0.1149 \\
    & Att & \textcolor{rhoPosBlue}{\textbf{0.2944}} & 0.1361 \\
    & Res & \textcolor{rhoNegRed}{\textbf{-0.2809}} & 0.1558 \\
    \rowcolor{sigLightGreen}  & MLP & \textcolor{rhoPosBlue}{\textbf{0.4046}} & 0.0363 \\
    \bottomrule
    \end{tabular} \\
    \end{tabular}
    }

    \caption{\textbf{Qwen -- LegalCase (metrics–performance drop correlations).}}
    \label{tab:layer_metric_ablation_correlations_qwen25_legalsumm}
    \end{table*}

\end{document}